%% file: PaperForReview.tex
\documentclass[10pt,twocolumn,letterpaper]{article}

\usepackage[pagenumbers]{cvpr} 

\usepackage{graphicx}
\usepackage{amsmath}
\usepackage{amssymb}
\usepackage{booktabs}

\usepackage{multirow}
\usepackage{diagbox}
\usepackage{pifont}
\newcommand{\cmark}{\ding{51}}%
\newcommand{\xmark}{\ding{55}}%

\usepackage{graphbox} 
\usepackage{colortbl}
\usepackage{xcolor}
\usepackage{collcell}
\definecolor{mygray}{gray}{.9}

\newcolumntype{a}{>{\columncolor{mygray}}l}
\newcolumntype{b}{>{\columncolor{white}}c}

\usepackage{color}
\definecolor{citecolor}{HTML}{0071bc}
\usepackage[pagebackref=true,breaklinks=true,colorlinks,citecolor=citecolor,bookmarks=false]{hyperref}

\usepackage[capitalize]{cleveref}
\crefname{section}{Sec.}{Secs.}
\Crefname{section}{Section}{Sections}
\Crefname{table}{Table}{Tables}
\crefname{table}{Tab.}{Tabs.}

\def\cvprPaperID{2460} 
\def\confName{CVPR}
\def\confYear{2023}

\usepackage{color}

\begin{document}

\title{Blind Video Deflickering by Neural Filtering with a Flawed Atlas}

\author{Chenyang Lei$^{1,2}$\thanks{Equal contribution} 
\quad  Xuanchi Ren$^{3,4}$\footnotemark[1]
\quad  Zhaoxiang Zhang$^1$ 
\quad  Qifeng Chen$^5$\\
$^1$CAIR, HKISI-CAS  \quad $^2$Princeton University  \quad $^3$University of Toronto \quad $^4$Vector Institute \quad $^5$HKUST \\
}

\twocolumn[{%
\renewcommand\twocolumn[1][]{#1}%
\maketitle
\begin{center}
\vspace{-1.5 em}
\renewcommand\arraystretch{0.5} 
\centering

\begin{tabular}{c@{\hspace{1mm}}c@{\hspace{0.5mm}}c@{\hspace{0.5mm}}c@{\hspace{0.5mm}}c}
\rotatebox{90}{\small \hspace{8mm} Input } &
\includegraphics[width=0.24\linewidth]{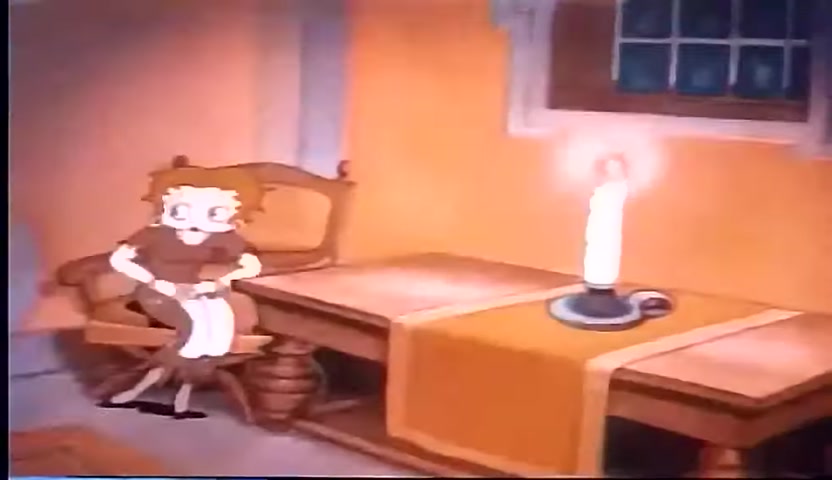}&
\includegraphics[width=0.24\linewidth]{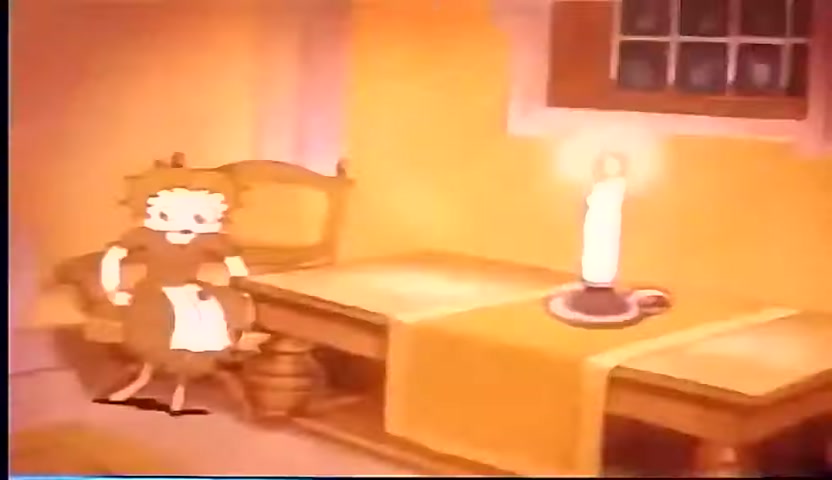}&
\includegraphics[width=0.24\linewidth]{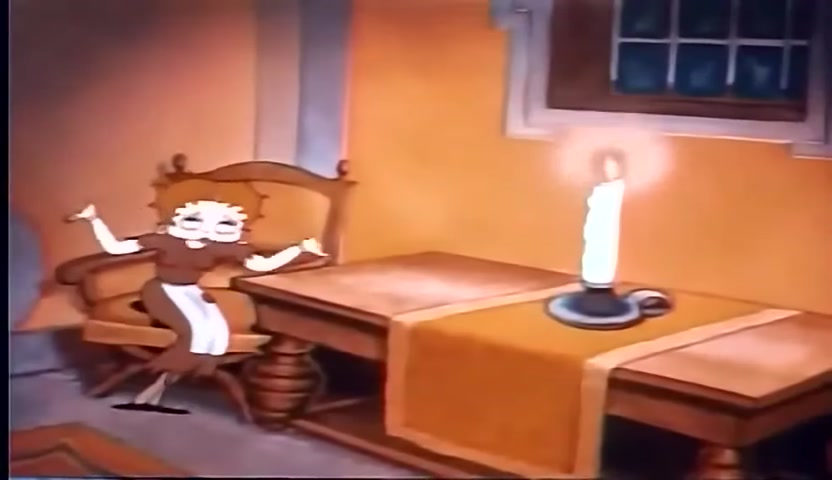}&
\includegraphics[width=0.24\linewidth]{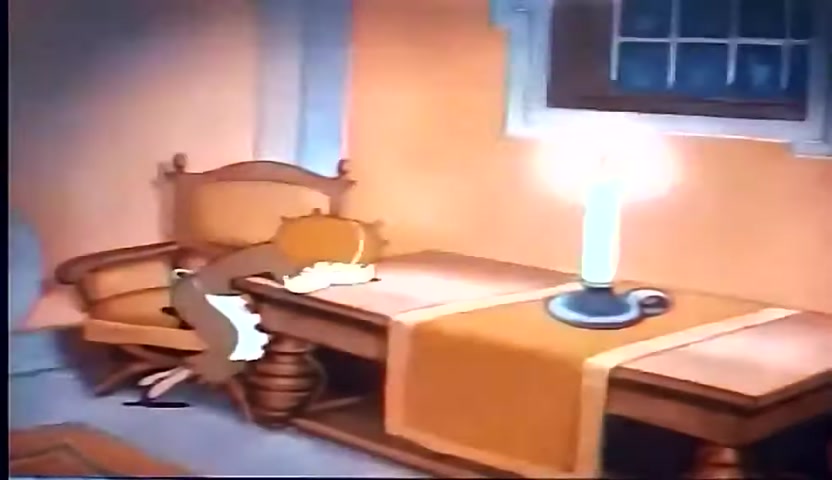}\\

\rotatebox{90}{\small \hspace{6mm} Output }&
\includegraphics[width=0.24\linewidth]{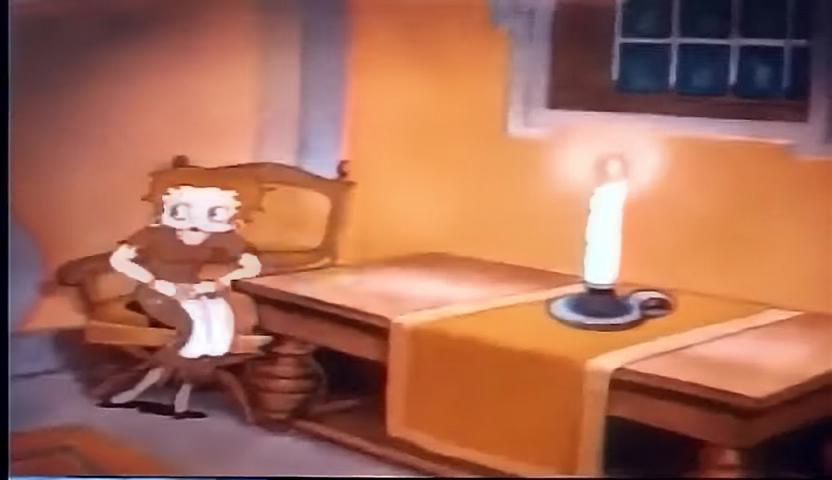}&
\includegraphics[width=0.24\linewidth]{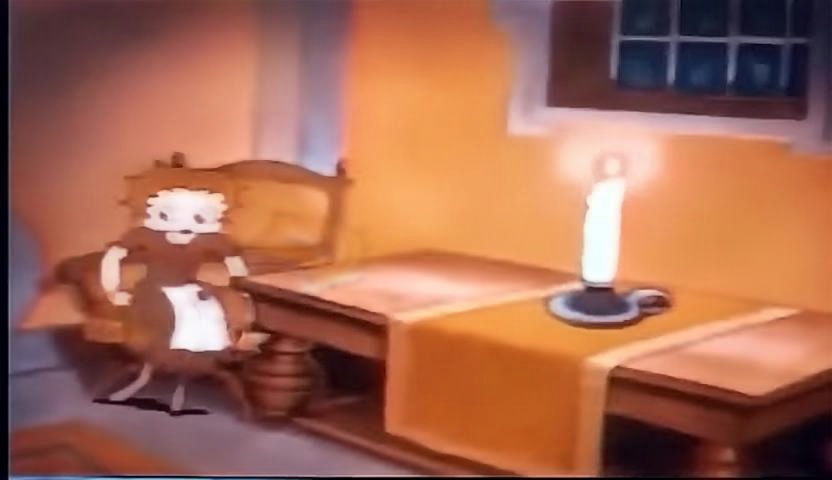}&
\includegraphics[width=0.24\linewidth]{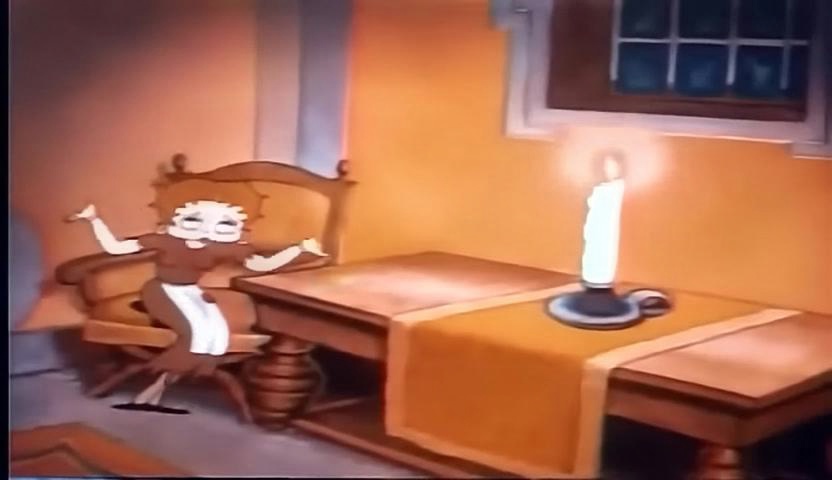}&
\includegraphics[width=0.24\linewidth]{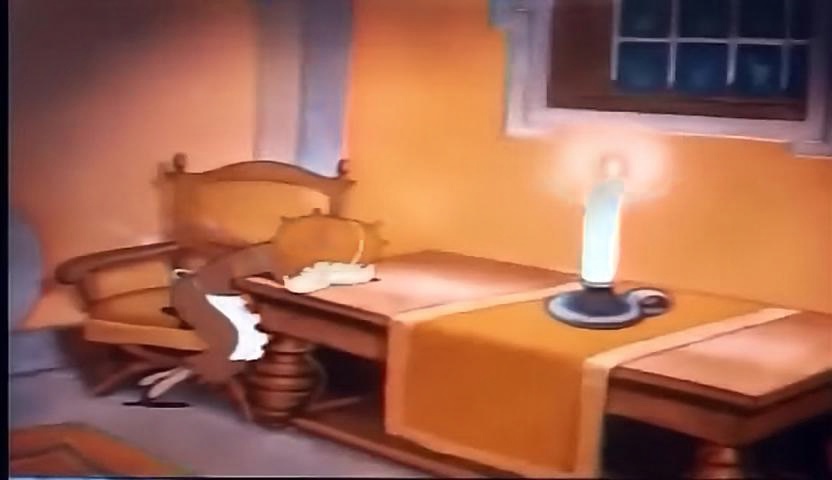}\\
\end{tabular}

\vspace{-0.5em}
\captionof{figure}{\textbf{Blind deflickering performance of our approach on \textit{Betty Boop} (1934).} Many videos can have flickering artifacts for various
reasons. Our approach takes only the input video and removes the flicker automatically without any extra guidance.}
\label{fig:teaser}
\end{center}
}]

{
  \renewcommand{\thefootnote}%
    {\fnsymbol{footnote}}
  \footnotetext[1]{Equal contribution.}
}

\begin{abstract}
Many videos contain flickering artifacts; common causes
of flicker include video processing algorithms, video generation
algorithms, and capturing videos under specific situations.
Prior work usually requires specific guidance such
as the flickering frequency, manual annotations, or extra
consistent videos to remove the flicker. In this work, we
propose a general flicker removal framework that only receives
a single flickering video as input without additional
guidance. Since it is blind to a specific flickering type or
guidance, we name this “blind deflickering.” The core of
our approach is utilizing the neural atlas in cooperation
with a neural filtering strategy. The neural atlas is a unified
representation for all frames in a video that provides
temporal consistency guidance but is flawed in many cases.
To this end, a neural network is trained to mimic a filter
to learn the consistent features (e.g., color, brightness) and
avoid introducing the artifacts in the atlas. To validate our
method, we construct a dataset that contains diverse real-world
flickering videos. Extensive experiments show that
our method achieves satisfying deflickering performance
and even outperforms baselines that use extra guidance on
a public benchmark. The source code is publicly available at \url{https://chenyanglei.github.io/deflicker}.
\end{abstract}

\input{sec1_introduction}
\input{sec2_relatedwork}
\input{sec3_method}

\input{sec4_dataset}

\input{sec5_experiment}

\section{Conclusion}
In this paper, we define a problem named \textit{blind deflickering} that can remove diverse flickering artifacts without knowing the specific flickering type and extra guidance. We propose the first dedicated approach for this task. The core of our approach is to adopt a neural atlas with a neural filtering strategy. The neural atlas concisely extracts all pixels in the videos and provides strong guidance to enforce long-term consistency, but it is flawed in many regions. We then use a neural network to filter the flaws of the atlas for satisfying performance. We conduct extensive experiments to evaluate the deflickering performance. Results show that our approach outperforms baselines significantly on different datasets, and controlled experiments validate the effectiveness of our key designs.

\section*{Acknowledgement}
\noindent This work was supported by the InnoHK program.


{\small
\bibliographystyle{ieee_fullname}
\bibliography{egbib}
}

\clearpage

\end{document}

%% file: sec1_introduction.tex
\begin{figure*}[t]
\centering
\begin{tabular}{@{}c@{\hspace{1mm}}c@{\hspace{1mm}}c@{\hspace{1mm}}c@{\hspace{1mm}}c@{\hspace{1mm}}c@{\hspace{1mm}}c@{\hspace{1mm}}c@{\hspace{1mm}}c@{}}
\rotatebox{90}{\hspace{1.2mm} \scriptsize{Frame 1}   }&

\includegraphics[height=0.069\linewidth]{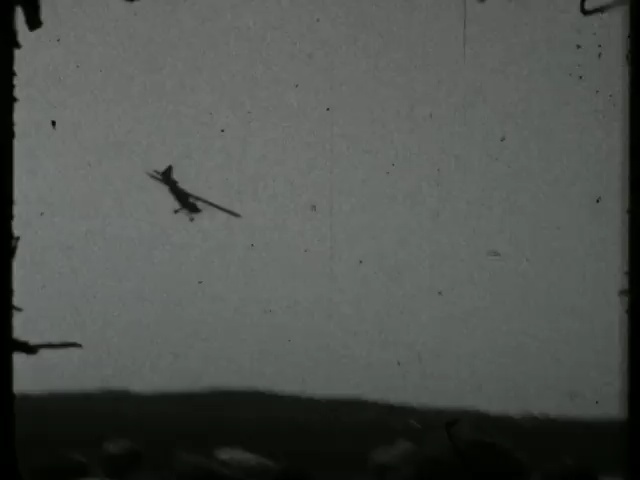}&
\includegraphics[height=0.069\linewidth]{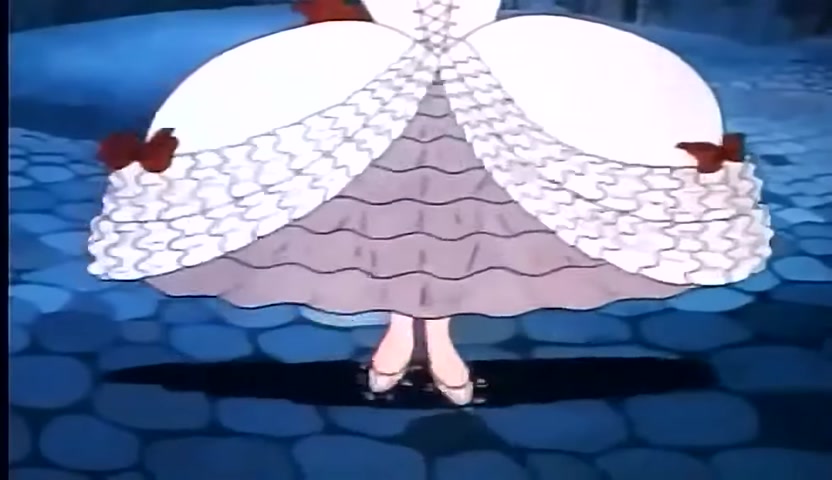}&
\includegraphics[height=0.069\linewidth]{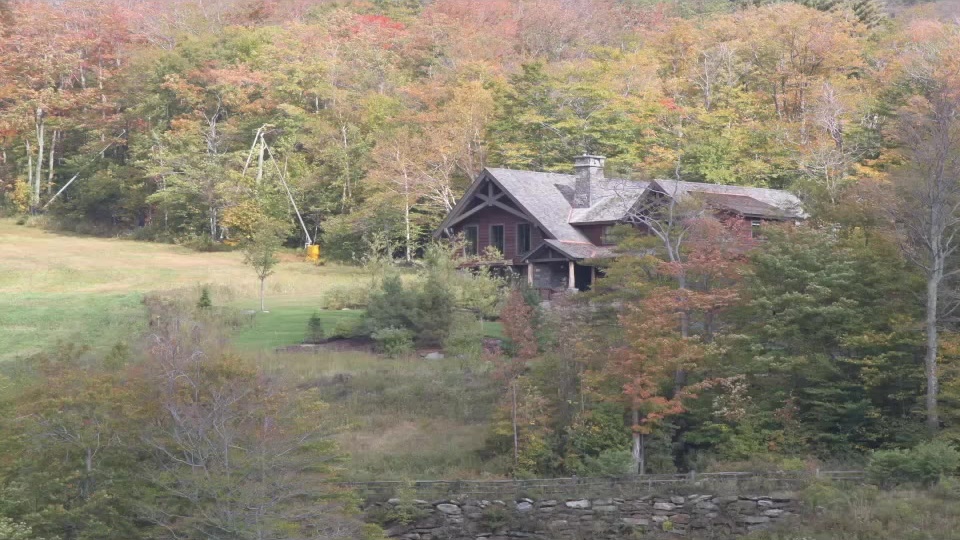}&
\includegraphics[height=0.069\linewidth]{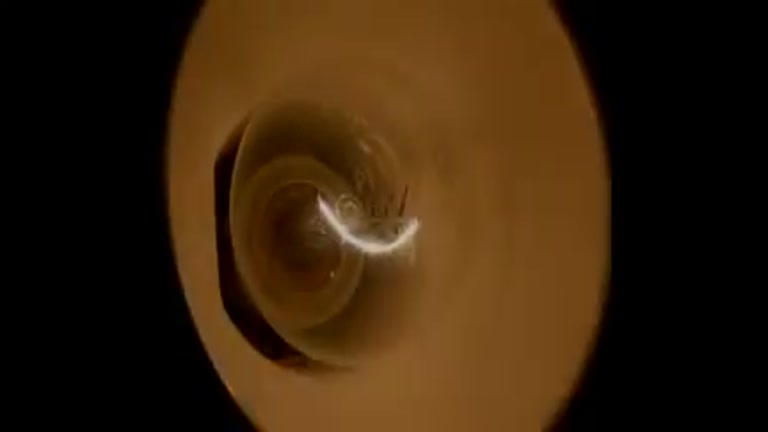}&
\includegraphics[height=0.069\linewidth]{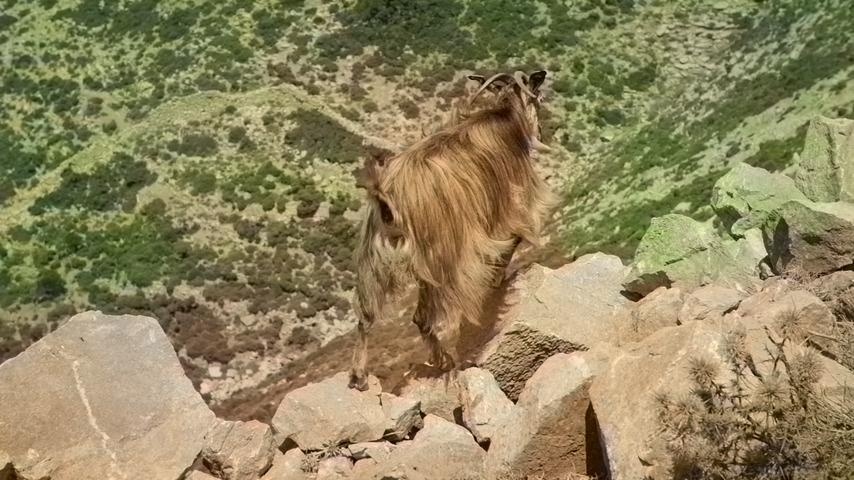}&
\includegraphics[height=0.069\linewidth]{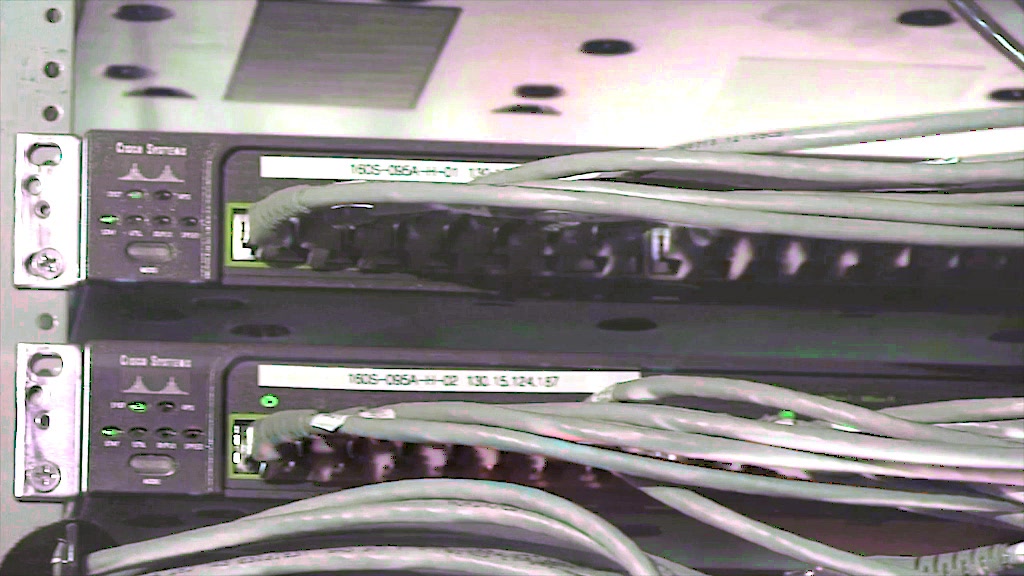}&
\includegraphics[height=0.069\linewidth]{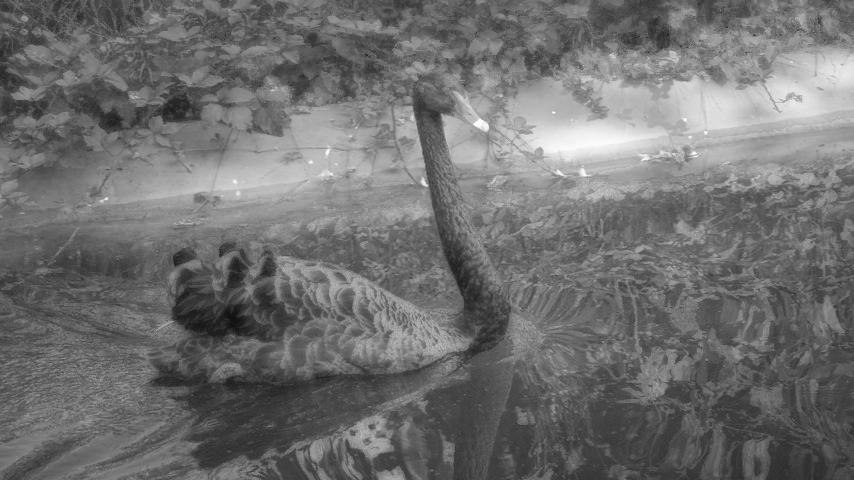}&
\includegraphics[height=0.069\linewidth, width=0.112\linewidth]{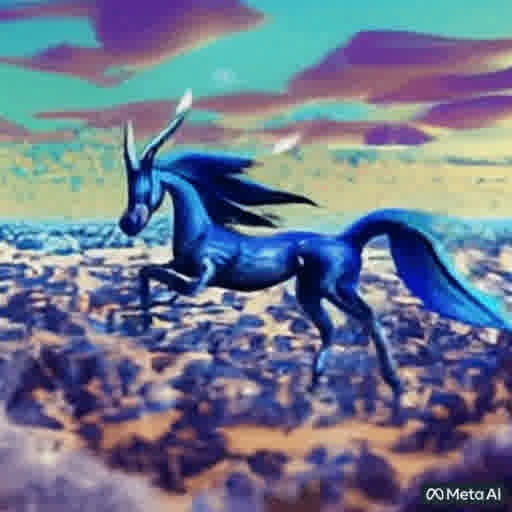}
\\
\rotatebox{90}{\hspace{1.2mm}  \scriptsize{Frame 2}  }&

\includegraphics[height=0.069\linewidth]{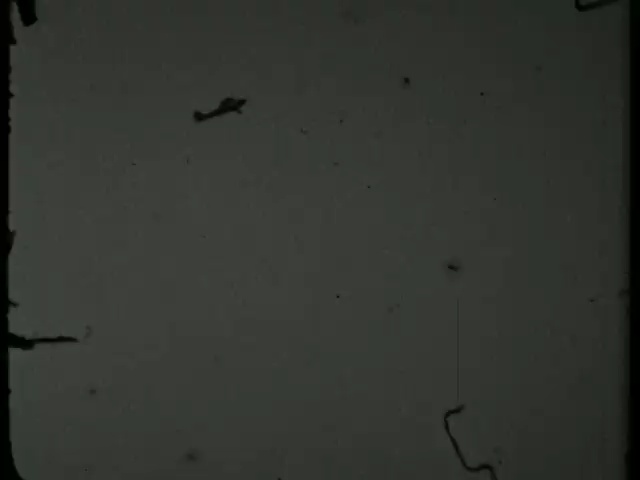}&
\includegraphics[height=0.069\linewidth]{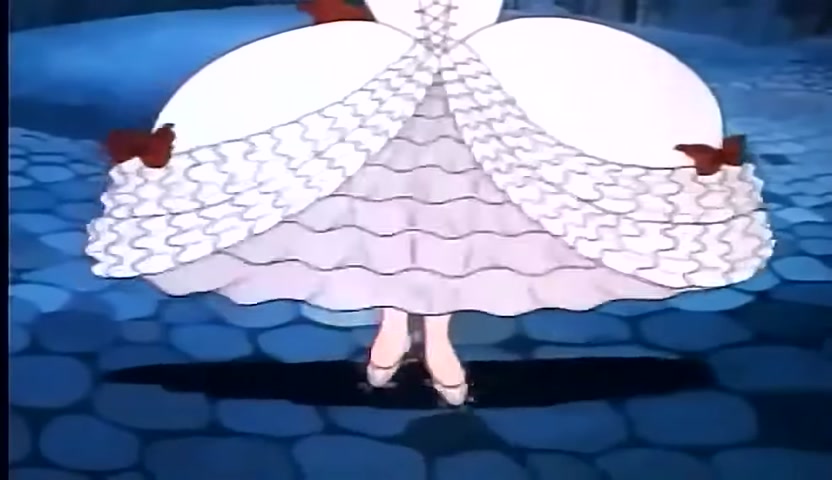}&
\includegraphics[height=0.069\linewidth]{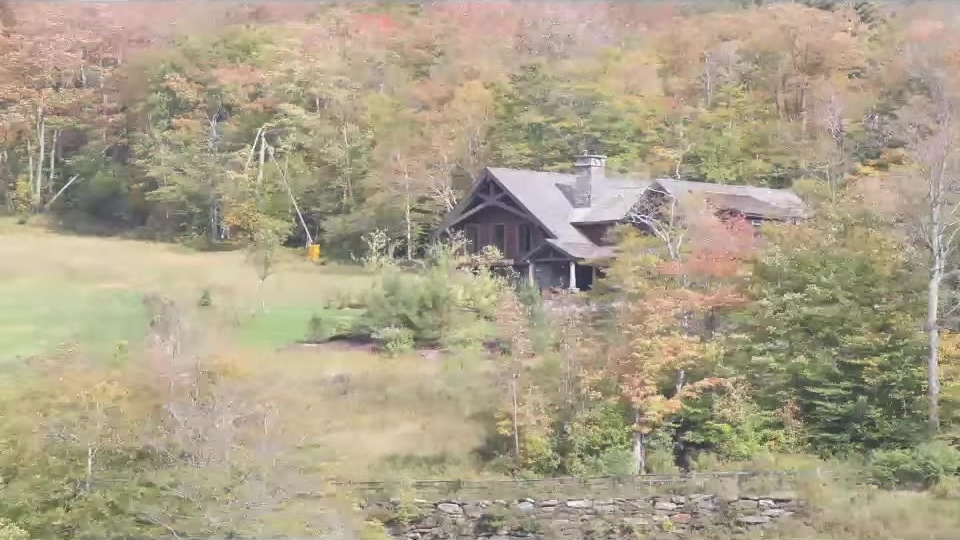}&
\includegraphics[height=0.069\linewidth]{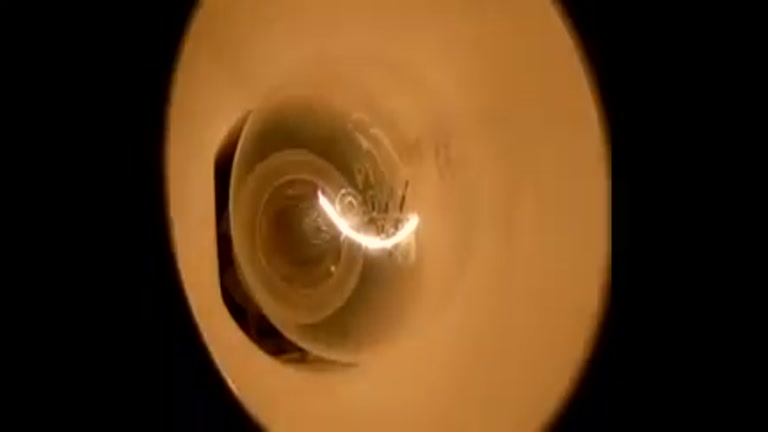}&
\includegraphics[height=0.069\linewidth]{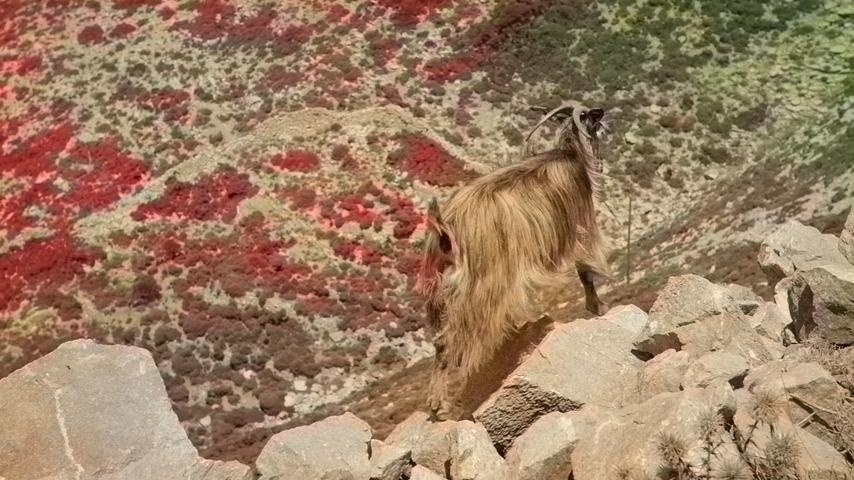}&
\includegraphics[height=0.069\linewidth]{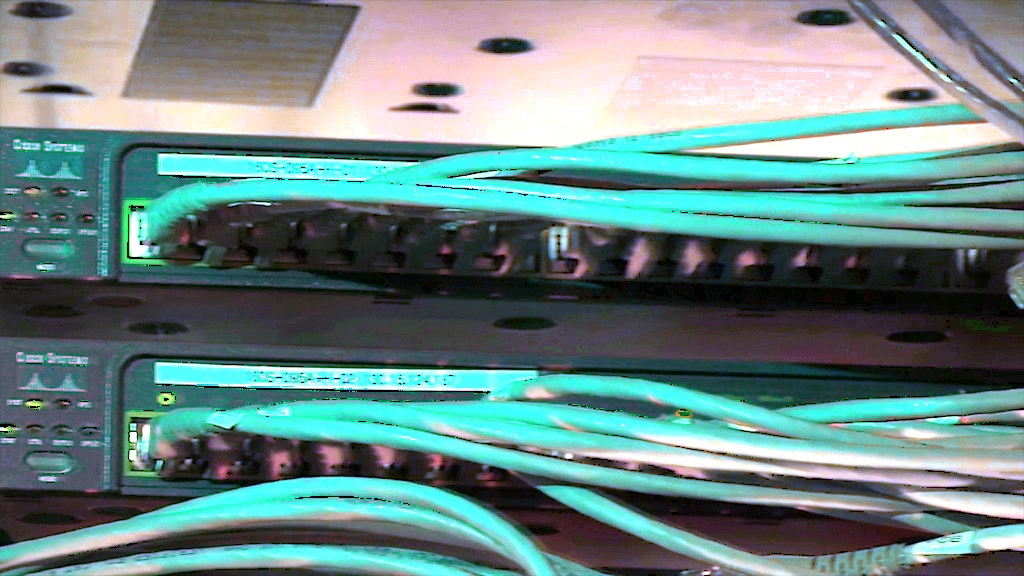}&
\includegraphics[height=0.069\linewidth]{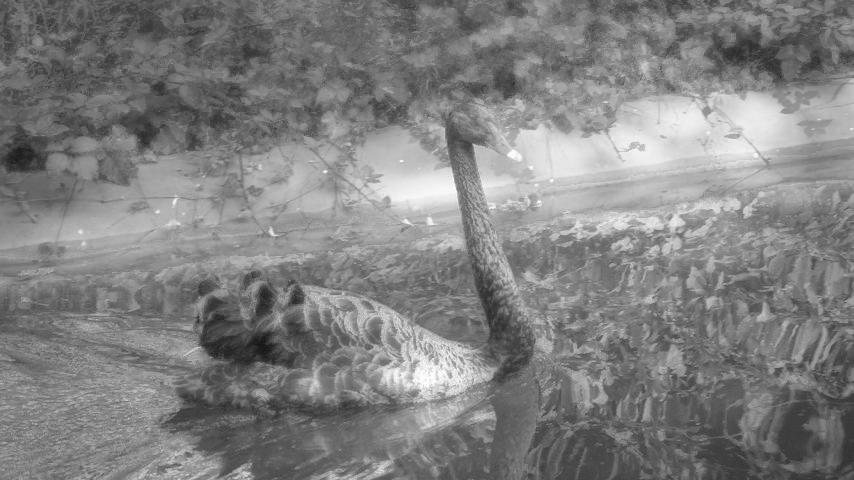}&
\includegraphics[height=0.069\linewidth, width=0.112\linewidth]{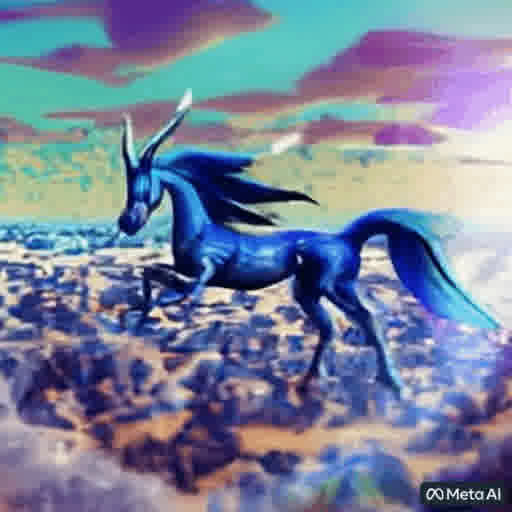}
\\
& Old movies&Old cartoons & Time-lapse& Slow-motion& Colorization&  White balance& Intrinsic&Text-to-video\\
\end{tabular}
\vspace{-0.5em}
\caption{\textbf{Videos with flickering artifacts.} Flickering artifacts exist in unprocessed videos, including old movies, old cartoons, time-lapse videos, and slow-motion videos. Besides, some processing algorithms~\cite{zhang2016colorful,he2010single,bell2014intrinsic} can introduce the flicker to temporally consistent unprocessed videos. Synthesized videos from video generations approaches~\cite{singer2022make,zhou2022magicvideo,DBLP:conf/cvpr/RenW22} also might be temporal inconsistent. \textit{Blind deflickering} aims to remove various types of flickering with only an unprocessed video as input.}
\label{fig:flicker_video}
\vspace{-1em}
\end{figure*}

\vspace{-3em}

\section{Introduction}
A high-quality video is usually temporally consistent, but many videos suffer from flickering artifacts for various reasons, as shown in Figure~\ref{fig:flicker_video}. For example, the brightness of old movies can be very unstable since some old cameras with low-quality hardware cannot set the exposure time of each frame to be the same~\cite{delon2010stabilization}. Besides, high-speed cameras with very short exposure time can capture the high-frequency (e.g., 60 Hz) changes of indoor lighting~\cite{kanj2017flicker}. Effective processing algorithms such as enhancement~\cite{marki2016bilateral,DBLP:journals/corr/abs-2012-05228}, colorization~\cite{zhang2016colorful,lei2019fully}, and style transfer~\cite{li2017universal} might bring flicker when applied to temporally consistent videos. Videos from video generations approaches~\cite{singer2022make,zhou2022magicvideo,DBLP:conf/cvpr/RenW22} also might contain flickering artifacts. Since temporally consistent videos are generally more visually pleasing, removing the flicker from videos is highly desirable in video processing~\cite{xie2020video,zhang2019deepcolor,chu2018temporally,chen2019seeing,claus2019videnn} and computational photography.

In this work, we are interested in a general approach for deflickering: (1) it is agnostic to the patterns or levels of flickering (e.g., old movies, high-speed cameras, processing artifacts), (2) it only takes a single flickering video and does not require other guidance (e.g., flickering types, extra consistent videos). That is to say, this model is blind to flickering types and guidance, and we name this task as \textit{blind deflickering}. Thanks to the blind property, blind deflickering has very wide applications.


Blind deflickering is very challenging since it is hard to enforce temporal consistency across the whole video without any extra guidance. Existing techniques usually design specific strategies for each flickering type with specific knowledge. For example, for slow-motion videos captured by high-speed cameras, prior work~\cite{kanj2017flicker} can analyze the lighting frequency. For videos processed by image processing algorithms, blind video temporal consistency~\cite{DBLP:conf/nips/dvp,lei2022deep} obtains long-term consistency by training on a temporally consistent unprocessed video. However, the flickering types or unprocessed videos are not always available, and existing flickering-specific algorithms cannot be applied in this case. One intuitive solution is to use the optical flow to track the correspondences. However, the optical flow from the flickering videos is not accurate, and the accumulated errors of optical flow are also increasing with the number of frames due to inaccurate estimation~\cite{bonneel2015blind}.

With two key observations and designs, we successfully propose the first approach for \textit{blind deflickering} that can remove various flickering artifacts without extra guidance or knowledge of flicker. First, we utilize a unified video representation named neural atlas~\cite{kasten2021layered} to solve the major challenge of solving long-term inconsistency. This neural atlas tracks all pixels in the video, and correspondences in different frames share the same pixel in this atlas. Hence, a sequence of temporally consistent frames can be obtained by sampling from the shared atlas. Secondly, while the frames from the shared atlas are consistent, the structures of images are flawed: the neural atlas cannot easily model dynamic objects with large motion; the optical flow used to construct the atlas is imperfect. Hence, we propose a neural filtering strategy to take the treasure and throw the trash from the flawed atlas. A neural network is trained to learn the invariant under two types of distortion, which mimics the artifacts in the atlas and the flicker in the video, respectively. At test time, this network works well as a filter to preserve the consistency property and block the artifacts from the flawed atlas.

We construct the first dataset containing various types of flickering videos to evaluate the performance of blind deflickering methods faithfully. Extensive experimental results show the effectiveness of our approach in handling different flicker. Our approach also outperforms blind video temporal consistency methods that use an extra input video as guidance on a public benchmark. 

Our contributions can be summarized as follows:\begin{itemize}
\setlength{\itemsep}{0pt}
\setlength{\parsep}{0pt}
\setlength{\parskip}{0pt}
\item We formulate the problem of blind deflickering and construct a deflickering dataset containing diverse flickering videos for further study.
\item We propose the first blind deflickering approach to remove diverse flicker. We introduce the neural atlas to the deflickering problem and design a dedicated strategy to filter the flawed atlas for satisfying deflickering performance.
\item Our method outperforms baselines on our dataset and even outperforms methods that use extra input videos on a public benchmark.

\end{itemize}

%% file: sec2_relatedwork.tex
\section{Related Work}

\noindent \textbf{Task-specific deflikcering.} Different strategies are designed for specific flickering types. Kanj et al.~\cite{kanj2017flicker} propose a strategy for high-speed cameras. Delon et al.~\cite{delon2010stabilization} present a method for local contrast correction, which can be utilized for old movies and biological film sequences. Xu et al.~\cite{DBLP:conf/eccv/XuAH22} focus on temporal flickering artifacts from GAN-based editing. Videos processed by a specific image-to-image translation algorithm~\cite{isola2017image,CycleGAN2017,he2010single,zhang2016colorful,lei2020polarized,bell2014intrinsic,li2017universal,ouyang2021neural} can suffer from flickering artifacts, and blind video temporal consistency~\cite{bonneel2015blind,lang2012practical,lai2018learning,DBLP:conf/nips/dvp,yao2017occlusion,lei2022deep} is designed to remove the flicker for these processed videos. These approaches are blind to a specific image processing algorithm. However, the temporal consistency of generated frames is guided by a temporal consistent unprocessed video. Bonneel et al.\cite{bonneel2015blind} compute the gradient of input frames as guidance. Lai et al.\cite{lai2018learning} input two consecutive input frames as guidance. Lei et al.~\cite{DBLP:conf/nips/dvp} directly learn the mapping function between input and processed frames. While these approaches achieve satisfying performance on many tasks, a temporally consistent video is not always available. For example, for many flickering videos such as old movies and synthesized videos from video generation methods~\cite{ho2022imagen,singer2022make,zhou2022magicvideo}, the original videos are temporal inconsistent. A concurrent preprint~\cite{abs-2206-03753} attempts to eliminate the need for unprocessed videos during inference. Nonetheless, it still relies on optical flow for local temporal consistency and does not study other types of flickering videos. Our approach has a wider application compared with blind temporal consistency. 

Also, some commercial software~\cite{revision,flicker_free} can be used for deflickering by integrating various task-specific deflickering approaches. However, these approaches require users to have a knowledge background of the flickering types. Our approach aims to remove this requirement so that more videos can be processed efficiently for most users.

\noindent \textbf{Video mosaics and neural atlas.}
Inheriting from panoramic image stitching~\cite{BrownL07}, video mosaicing is a technique that organizes video data into a compact mosaic representation, especially for dynamic scenes.
It supports various applications, including video compression~\cite{IraniAH95}, video indexing~\cite{IraniA98}, video texture~\cite{AgarwalaZPACCSS05}, 2D to 3D conversion~\cite{RiberaCKLN12, SchnyderWS11}, and video editing~\cite{Rav-AchaKRF08}. 
Building video mosaics based on homography warping often fails to depict motions. To handle the dynamic contents, researchers compose foreground and background regions by spatio-temporal masks~\cite{CorreaM10}, or blend the video into a multi-scale tapestry in a patch-based manner~\cite{BarnesGSF10}. 
However, these approaches heavily rely on image appearance information and thus are sensitive to lighting changes and flicker. 

Recently, aiming for consistent video editing, Kasten et al.~\cite{kasten2021layered} propose Neural Layered Atlas (NLA), which decomposes a video into a set of atlases by learning mapping networks between atlases and video frames.
Editing on the atlas and then reconstructing frames from the atlas can achieve consistent video editing.
Follow-up work validates its power on text-driven video stylization~\cite{abs-2206-12396,text2live} and face video editing~\cite{LiuCLLJFG22}. 
Directly adopting NLA to blind deflickering tasks is not trivial and is mainly limited by two-fold: 
\textit{(i)} its performance on the complex scene is still not satisfying with notable artifacts.
\textit{(ii)} it requires segmentation masks as guidance for decomposing dynamic objects, and each dynamic object requires an additional mapping network. 
To facilitate automatic deflickering, we use a single-layered atlas without the need for segmentation masks by designing an effective neural filtering strategy for the flawed atlas. Our proposed strategy is also compatible with other atlas generation techniques.

\begin{figure*}[t]
\centering
\begin{tabular}{@{}c@{}}
\includegraphics[width=1.0\linewidth]{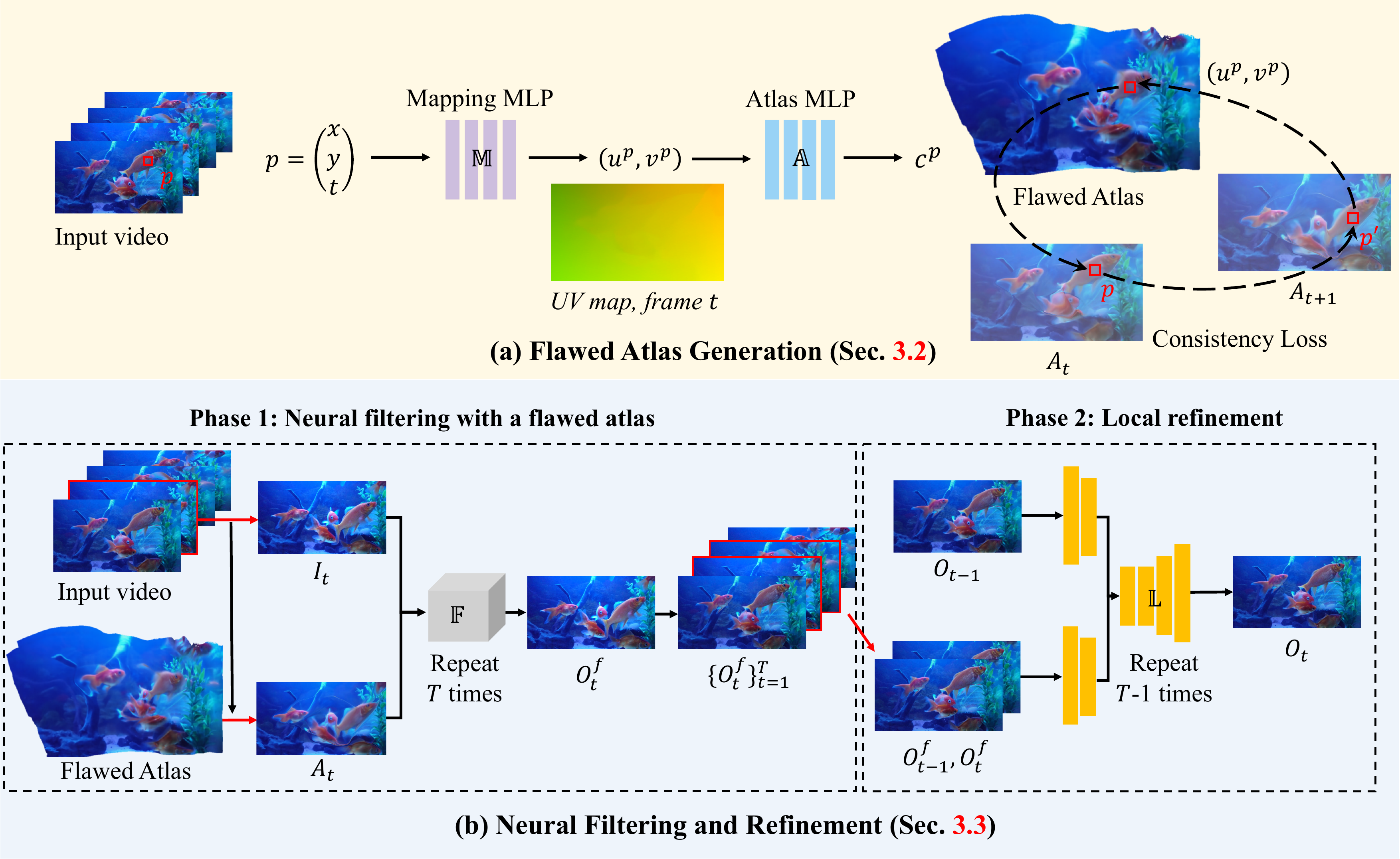}
\end{tabular}
\vspace{-0.5em}
\caption{\textbf{The framework of our approach}. We first generate an atlas as a unified representation of the whole video, providing consistent guidance for deflickering. Since the atlas is flawed, we then propose a neural filtering strategy to filter the flaws. }
\label{fig:framework}
\vspace{-1.2em}
\end{figure*}

\noindent \textbf{Implicit image/video representations.}
With the success of using the multi-layer perceptron (MLP) as a continuous implicit representation for 3D geometry~\cite{MeschederONNG19,nerf,ParkFSNL19}, such representation has gained popularity for representing images and videos~\cite{DBLP:conf/nips/SitzmannMBLW20, NERV, LiNSW21, TancikSMFRSRBN20}. Follow-up work extends these models to various tasks, such as image super-resolution~\cite{ChenL021}, video decomposition~\cite{DBLP:conf/cvpr/YeL0KS22}, and semantic segmentation~\cite{DBLP:conf/eccv/HuCXBCPW22}. 
In our work, we follow \cite{kasten2021layered} to employ a coordinate-based MLP to represent the neural atlases. 

%% file: sec3_method.tex
\section{Method}
\label{sec:background}
\subsection{Overview}
\label{subsec:overview}

Let $\{I_t\}_{t=1}^T$ be the input video sequences with flickering artifacts where $T$ is the number of video frames. Our approach aims to remove the flickering artifacts to generate a temporally consistent video $\{O_t\}_{t=1}^T$. Flicker denotes a type of temporal inconsistency that correspondences in different frames share different features (e.g., color, brightness). The flicker can be either global or local in the spatial dimension, either long-term or short-term in the temporal dimension.

Figure~\ref{fig:framework} shows the framework of our approach. We tackle the problem of blind deflickering through the following key designs: 
$(i)$ We first propose to use a single \textit{neural atlas (Section~\ref{subsec:atlas})} for the deflickering task. 
$(ii)$ We design a \textit{neural filtering (Section~\ref{subsec:filter})} strategy for the neural atlas as the single atlas is inevitably flawed.


\subsection{Flawed Atlas Generation}
\label{subsec:atlas}



\noindent\textbf{Motivation.}
A good blind deflickering model should have the capacity to track correspondences across all the video frames. Most architectures in video processing can only take a small number of frames as input, resulting in a limited receptive field that is insufficient for long-term consistency. To this end, we introduce neural atlases~\cite{kasten2021layered} to the deflickering task as we observe it perfectly fits this task. A neural atlas is a unified and concise representation of all the pixels in a video. As shown in Figure~\ref{fig:framework}(a), let $p =(x,y,t) \in \mathbb{R}^3 $ be a pixel that locates at $(x,y)$ in frame $I_t$. Each pixel $p$ is fed into a mapping network $\mathbb M$ that predicts a 2D coordinate $(u^p,v^p)$ representing the corresponding position of the pixel in the atlas. Ideally, the correspondences in different frames should share a pixel in the atlas, even if the color of the input pixel is different. That is to say, temporal consistency is ensured.


\noindent\textbf{Training.} Figure~\ref{fig:framework}(a) shows the pipeline to generate the atlas. For each pixel $p$, we have:
\begin{align}
(u^p,v^p)&=\mathbb M(p), \\
c^p & = \mathbb{A}(\phi(u^p),\phi(v^p)).
\end{align}
The 2D coordinate $(u^p,v^p)$ is fed into an atlas network $\mathbb A$ with positional encoding function $\phi(\cdot)$ to estimate the RGB color $c^p$ for the pixel. The mapping network $\mathbb M$ and the atlas network $\mathbb A$ are trained jointly by optimizing the loss between the original RGB color and the predicted color $c^p$. 
Besides the reconstruction term, a consistency loss is also employed to encourage the corresponding pixels in different frames to be mapped to the same atlas position. We follow the implementation of loss functions in \cite{kasten2021layered}. 
After training the networks $\mathbb M$ and $\mathbb A$, we can reconstruct the videos $\{A_t\}_{t=1}^T$ by providing all the coordinates of pixels in the whole video. The reconstructed atlas-based video $\{A_t\}_{t=1}^T$ is temporally consistent as all the pixels are mapped from a single atlas. In this work, we utilize the temporal consistency property of this video for blind deflickering.

Considering the trade-off between performance and efficiency, we only use a single-layer atlas to represent the whole video, although using two layers (background layer and foreground layer)  or multiple layers of atlases might slightly improve the performance. First, in practice, we notice the number of layers is quite different, which varies from a single layer to multiple layers (more than two), making it challenging to apply them to diverse videos automatically. Besides, we notice that artifacts and distortion are inevitable for many scenes, as discussed in \cite{kasten2021layered}. We present how to handle the artifacts in the flawed atlas with the following network designs in Secion~\ref{subsec:filter}.

\subsection{Neural Filtering and Refinement}
\label{subsec:filter}

\noindent \textbf{Motivation.} 
The neural atlas contains not only the treasure but also the trash. In Section~\ref{subsec:atlas}, we argue that an atlas is a strong cue for blind deflickering since it can provide consistent guidance across the whole video. However, the reconstructed frames from the atlas are flawed. First, as analyzed in NLA~\cite{kasten2021layered}, it cannot perform well when the object is moving rapidly, or multiple layers might be required for multiple objects. For blind deflickering, we need to remove the flicker and avoid introducing new artifacts. Secondly, the optical flow obtained by the flickering video is not accurate, which leads to more flaws in the atlas. At last, there are still some natural changes, such as shadow changes, and we hope to preserve the patterns. 

Hence, we design a \textit{neural filtering} strategy that utilizes the promising temporal consistency property of the atlas-based video and prevents the artifacts from destroying our output videos.

\begin{figure}[t]
\centering
\begin{tabular}{@{}c@{}}
\includegraphics[width=1.0\linewidth]{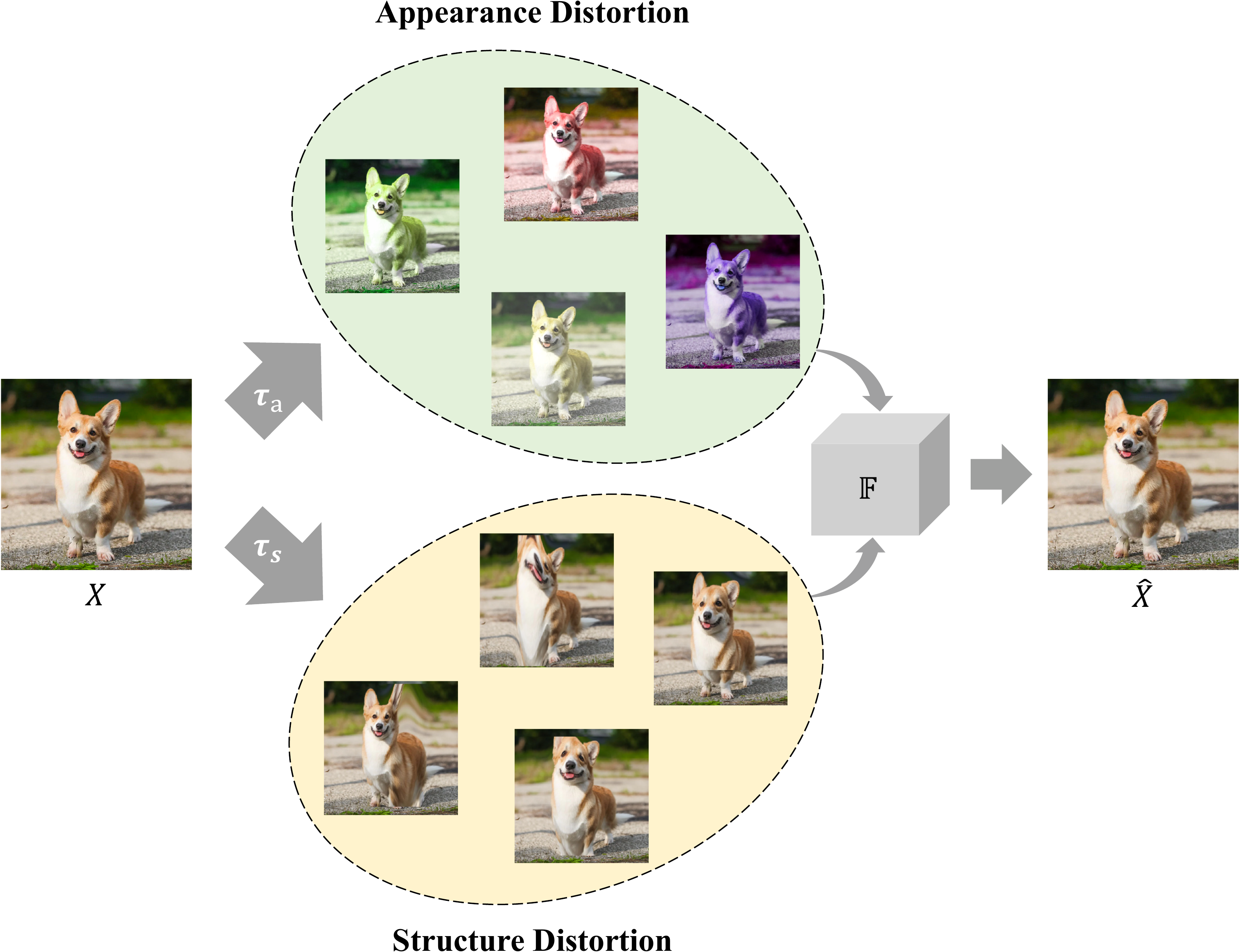}
\end{tabular}
\vspace{-0.5em}
\caption{\textbf{Training pipeline of our neural filtering strategy.} We apply two transformations to a clean image $X$ for mimicking the flickering input frame and the flawed atlas frame.}
\label{fig:NeuralFilter}
\vspace{-1em}
\end{figure}

\noindent \textbf{Training Strategy.} 
Figure~\ref{fig:framework}(b) shows the framework to use the atlas. Given an input video $\{I_t\}_{t=1}^T$ and an atlas $A$, in every iteration, we get one frame $I_t$ from the video and input it to the filter network $\mathbb F$:
\begin{align}
    O_t^f = {\mathbb F}(I_t, A_t),
\end{align}
where $A_t$ is obtained by fetching pixels from the shared atlas $A$ with the coordinates of $I_t$. 

We design a dedicated training strategy for the flawed atlas, as shown in Figure~\ref{fig:NeuralFilter}. We train the network using only a single image $X$ instead of consecutive frames. In the training time, we apply a transformation $\tau_a(\cdot)$ to distort the appearance, including color, saturation, and brightness of the image, which mimics the flickering pattern in $I_t$. We apply another transformation $\tau_s(\cdot)$ to distort the structures of the image, mimicking the distortion of a flawed atlas-based frame in $A_t$.  
At last, the network is trained by minimizing the L2 loss function $\mathcal{L}$ between prediction ${\mathbb F}(\tau_a(X), \tau_s(X))$ and the clean ground truth $X$ (i.e., the image before augmentation):
\begin{align}
    \mathcal{L} = ||{\mathbb F}(\tau_a(X), \tau_s(X);\theta_{\mathbb F}) - X||_2^2,
\end{align}
where $\theta_{\mathbb F}$ is the parameters of filtering network $\mathbb F$. 

The network tends to learn the invariant part from two distorted views respectively. Specifically, ${\mathbb F}$ learns the structures from the input frame $I_t$ and the appearance (e.g., brightness, color) from the atlas frame $A_t$ as they are invariant to the structure distortion $\tau_s$. At the same time, the distortion of $\tau_s(X)$ would not be passed through the network $\mathbb F$. With this strategy, we achieve the goal of neural filtering with the flawed atlas.

Note that while this network ${\mathbb F}$ only receives one frame, long-term consistency can be enforced since the temporal information is encoded in the atlas-based frame $A_t$.

\noindent \textbf{Local refinement.}
\label{subsec:local}
The video frames $\{O_t^f\}_{t=1}^T$ are consistent with each other globally in both the short term and the long term. However, it might contain local flicker due to the misalignment between input and atlas-based frames. Hence, we use an extra local deflickering network to refine the results further. Prior work has shown that local flicker can be well addressed by a flow-based regularization. We thus choose a lightweight pipeline~\cite{lai2018learning} with modification. As shown in Figure~\ref{fig:framework}(b), we predict the output frame $O_t$ by providing two consecutive frames $O_t^f, O_{t-1}^f$ and previous output $O_{t-1}$ to our local refinement network ${\mathbb L}$. Two consecutive frames are firstly followed by a few convolution layers and then fused with the $O_{t-1}$. The local flickering network is trained with a simple temporal consistency loss $\mathcal L_{local} $ to remove local flickering artifacts:
\begin{align}
    \mathcal{L}_{local}(O_t,O_{t-1}) &=  ||M_{t,t-1} \odot (O_t - \hat{O}_{t-1})||_1 ,
\end{align}
where $\hat{O}_{t-1}$ is obtained by warping the $O_{t-1}$ with the optical flow from frame $t$ to frame $t-1$. $M_{t,t-1}$ is the corresponding occlusion mask. For the frames without local artifacts, the output should $O_t$ be the same as $O_t^f$. Hence, we also provide a reconstruction loss by minimizing the distance between $O_t$ and $O_t^f$ to regularize the quality.


\noindent \textbf{Implementation details.}
The network ${\mathbb F}$ is trained on the MS-COCO dataset~\cite{coco} as we only need images for training. We train it for $20$ epochs with a batch size of $8$. For the network ${\mathbb L}$, we train it on the processed DAVIS dataset~\cite{davis,lai2018learning} for $50$ epochs with a batch size of $8$. We adopt the Adam optimizer and set the learning rate to 0.0001.

\begin{table*}[t]
\centering
\begin{tabular}{cc}%
\hfill
\begin{minipage}{0.62\textwidth}
\centering
\renewcommand{\arraystretch}{1.1}
\begin{tabular}{lccccc}
\toprule
 \cellcolor{white}Video type    & \multicolumn{3}{c}{$E_{warp} \downarrow$} & \multicolumn{2}{c}{ PSNR $\uparrow$}\\
 \cellcolor{white} & {Raw video} & ConvLSTM &  {Ours} & ConvLSTM &  {Ours} \\
 \hline
 Synthetic & 0.163 & 0.148 & \textbf{0.088} & 21.84 & \textbf{26.46} \\
  \qquad \footnotesize{-- $W = 1$}& 0.199 & 0.168 & \textbf{0.091} & 19.84 & \textbf{27.75} \\
 \qquad \footnotesize{-- $W = 3$}& 0.158 & 0.151 & \textbf{0.086} & 21.71 & \textbf{26.40} \\
 \qquad \footnotesize{-- $W = 10$} & 0.132 & 0.124 & \textbf{0.086} &  23.98 & \textbf{25.21} \\
 \hline
 Processed & 0.128 & 0.118 & \textbf{0.094} & -- & --\\

 \bottomrule
\end{tabular}
\caption*{(a) Quantitative comparison} 

\label{fig:plogp}
\end{minipage}
&
\begin{minipage}{0.33\textwidth}
\centering
\renewcommand{\arraystretch}{1.1}
\begin{tabular}{lbb}
\toprule
\cellcolor{white}Video type & \multicolumn{2}{c}{Preference rate} \\
\cellcolor{white} & ConvLSTM & Ours  \\
\hline
{Old movies} & 38.0\% & \textbf{62.0\%} \\
{Old cartoons} & 33.6\% & \textbf{66.4\%} \\
Time-lapse & 31.9\% & \textbf{68.1\%} \\
Slow-motion & 29.0\% & \textbf{71.0\%} \\
\hline
Average & 33.7\% & \textbf{66.3\%} \\
\bottomrule
\end{tabular}
\caption*{(b) User study} 
\end{minipage}%
\end{tabular}
\caption{\textbf{Comparison to baselines.} We provide the (a) quantitative comparison for processed and synthetic videos since we have the high-quality optical flow for computing the evaluation metric. The warping errors of our approach are much smaller than the baseline, and our results are more similar to the ground truth on synthetic data, according to the PSNR. For the other real-world videos that cannot provide high-quality optical flow, we provide the (b) user study results for comparison. Our results are preferred by most users. } 
\label{table:comparison_ourdata}
\end{table*}

\begin{figure*}[t]
\centering
\begin{tabular}{@{}c@{\hspace{1mm}}c@{\hspace{1mm}}c@{\hspace{1mm}}c@{\hspace{1mm}}c@{\hspace{1mm}}c@{}}

\rotatebox{90}{\small \hspace{2mm}  }&
\includegraphics[width=0.230\linewidth]{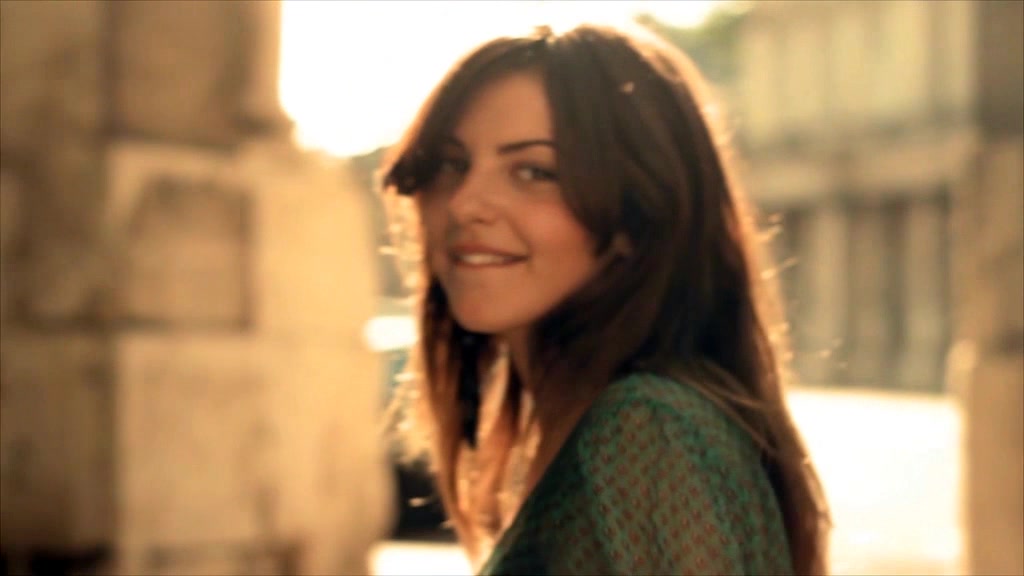}&
\includegraphics[width=0.230\linewidth]{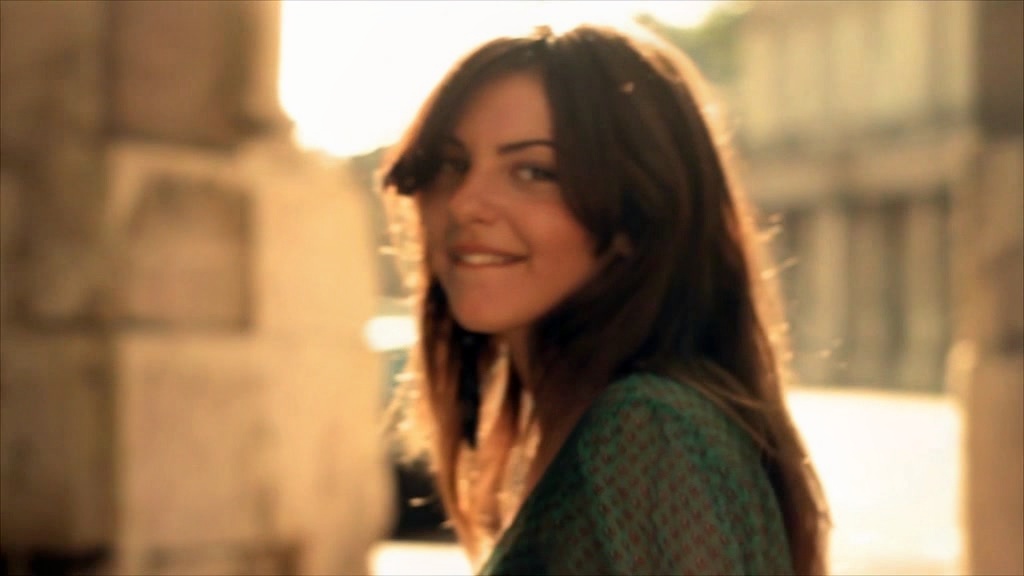}&
\includegraphics[width=0.230\linewidth]{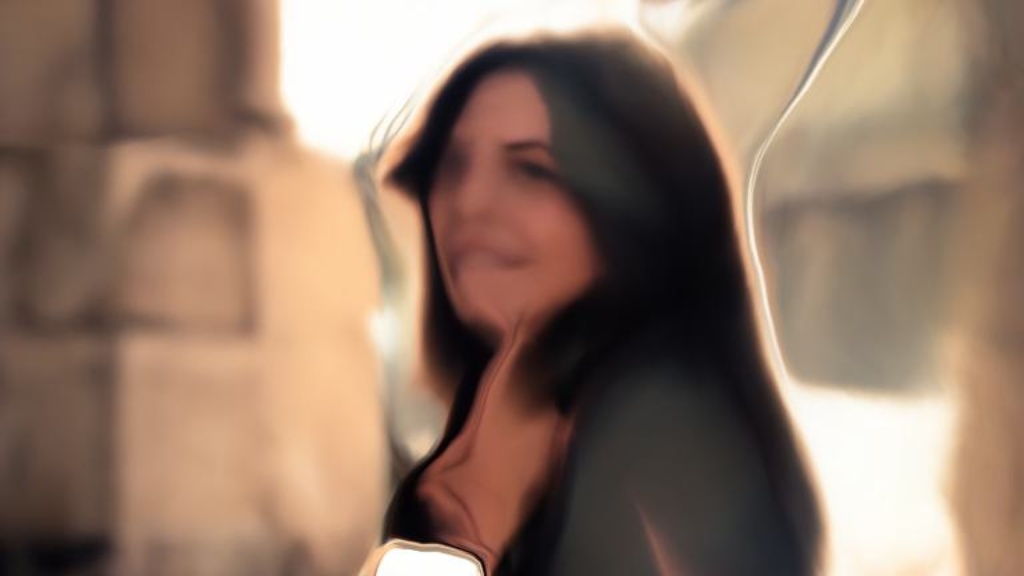}&
\includegraphics[width=0.230\linewidth]{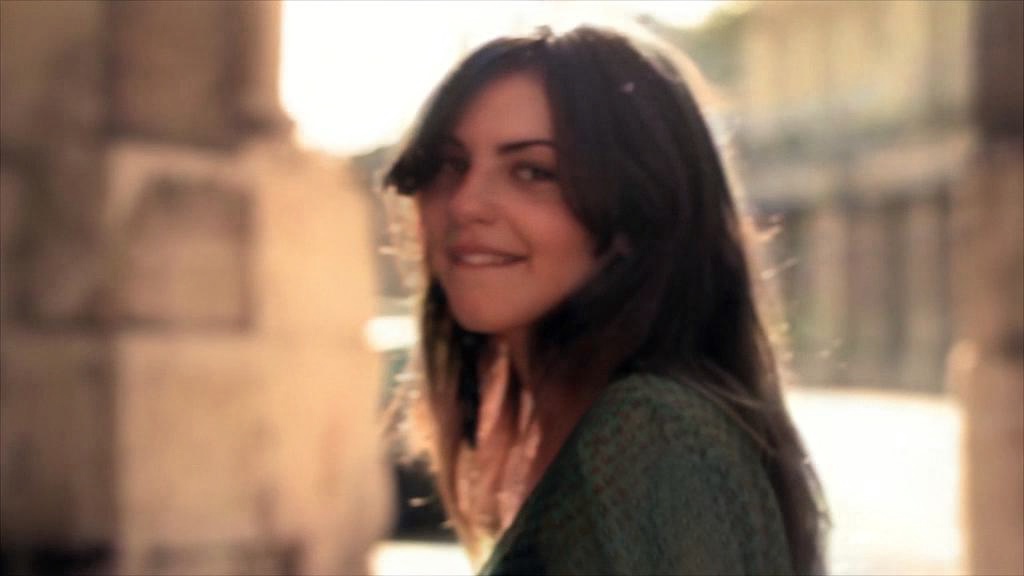}\\
\rotatebox{90}{\small \hspace{2mm}  }&
\includegraphics[width=0.230\linewidth]{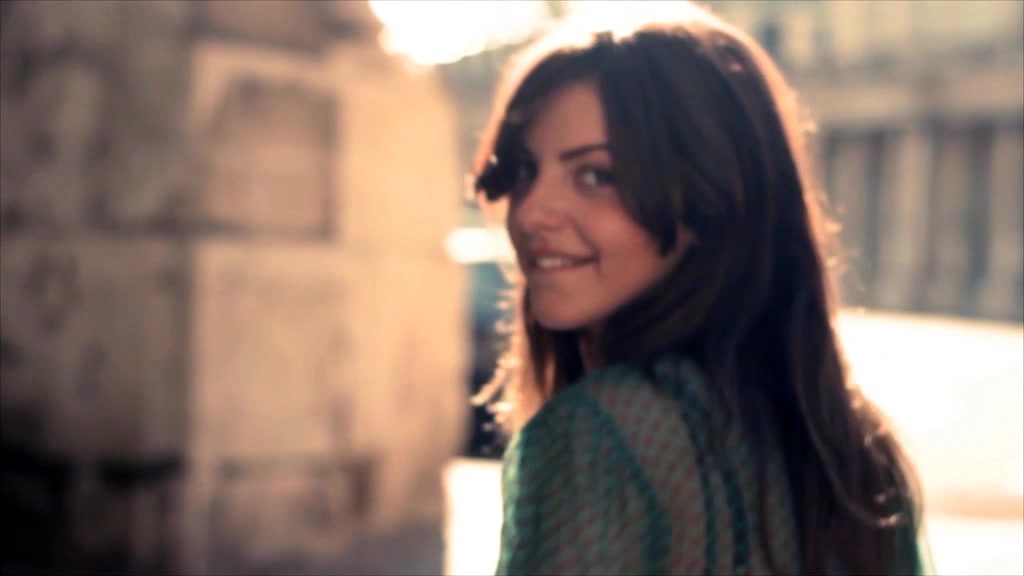}&
\includegraphics[width=0.230\linewidth]{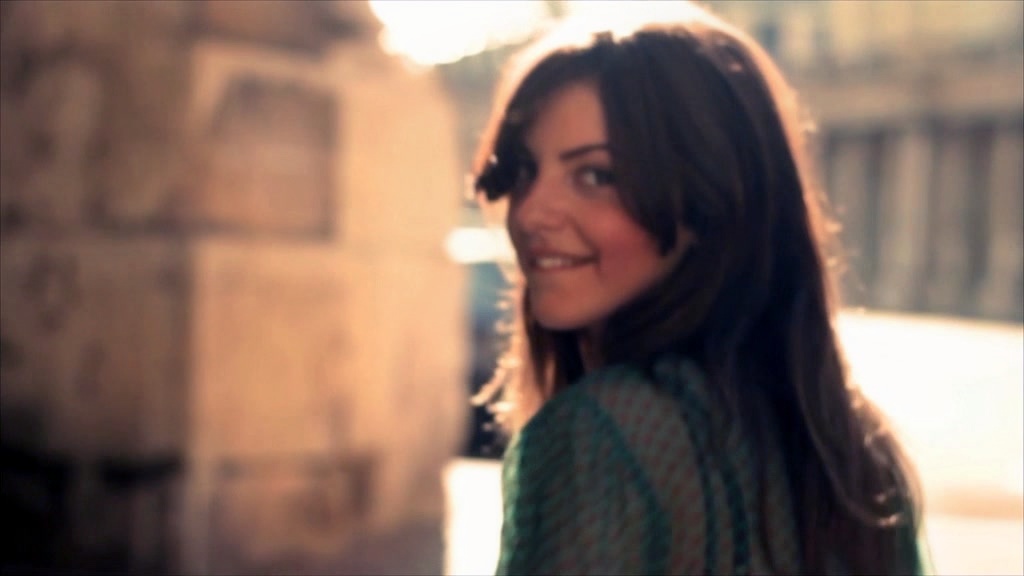}&
\includegraphics[width=0.230\linewidth]{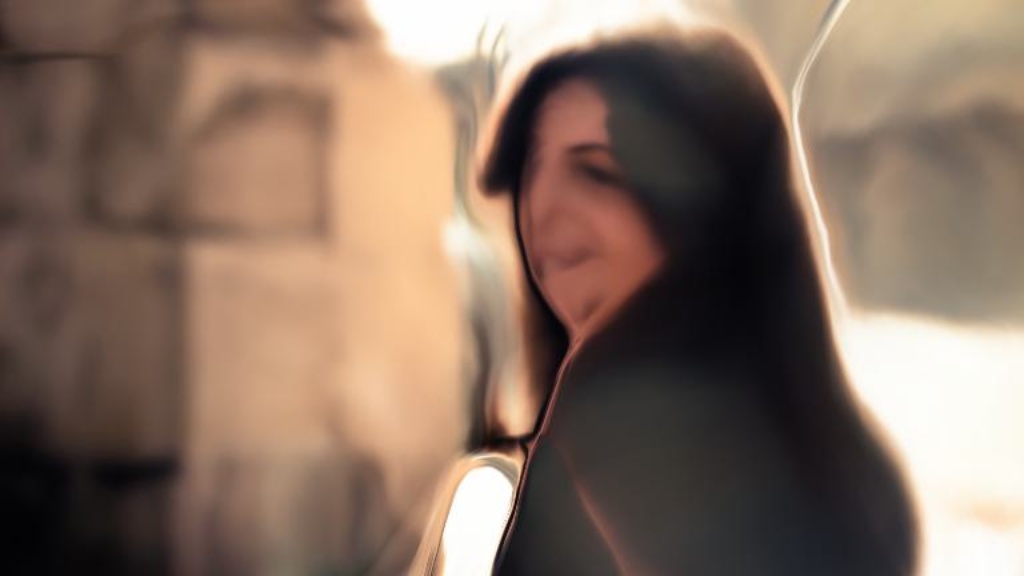}&
\includegraphics[width=0.230\linewidth]{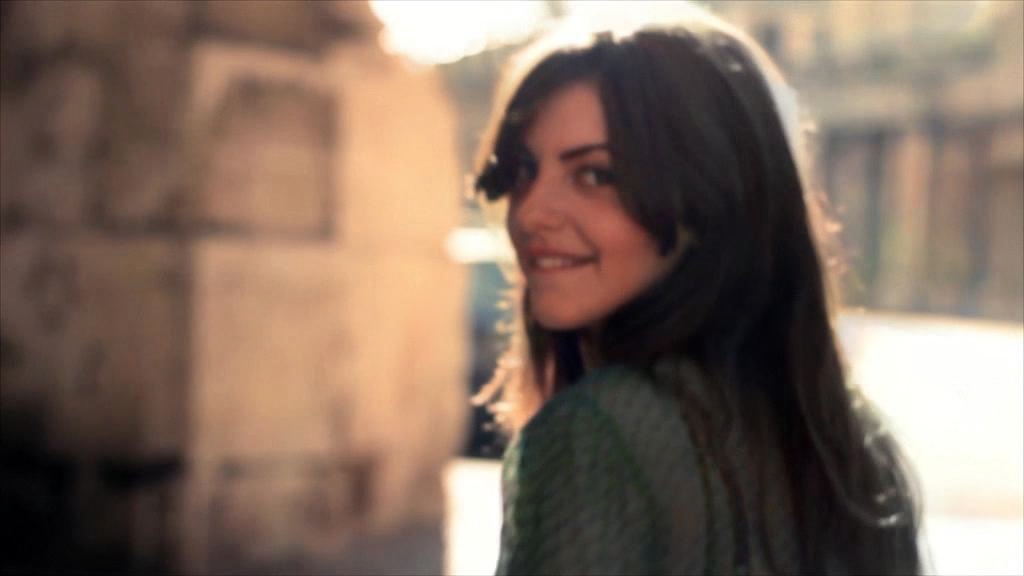}\\

\rotatebox{90}{\small \hspace{2mm}  }&
\includegraphics[width=0.230\linewidth]{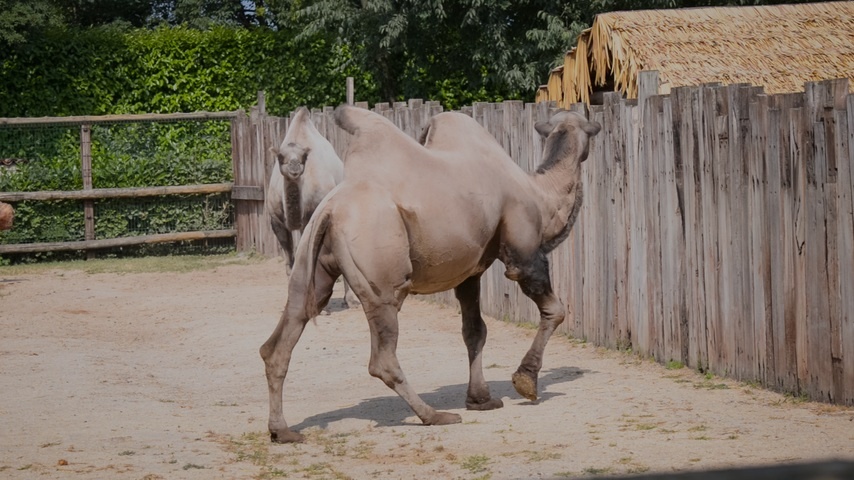}&
\includegraphics[width=0.230\linewidth]
{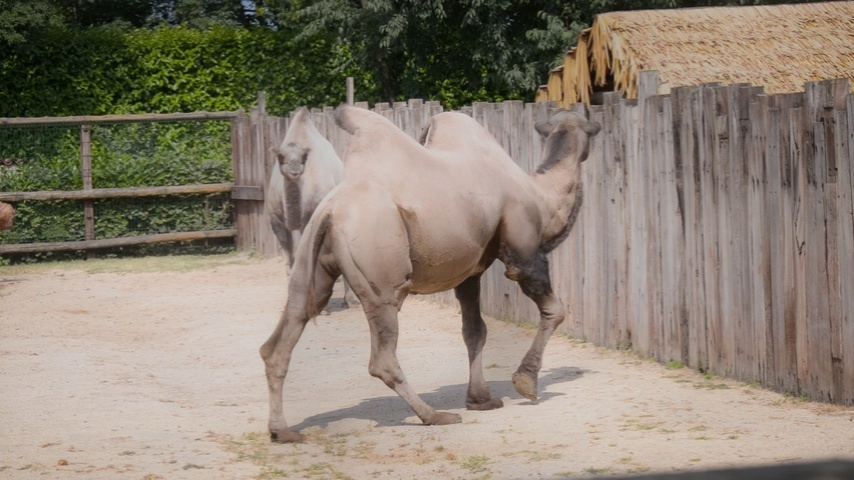}&
\includegraphics[width=0.230\linewidth]{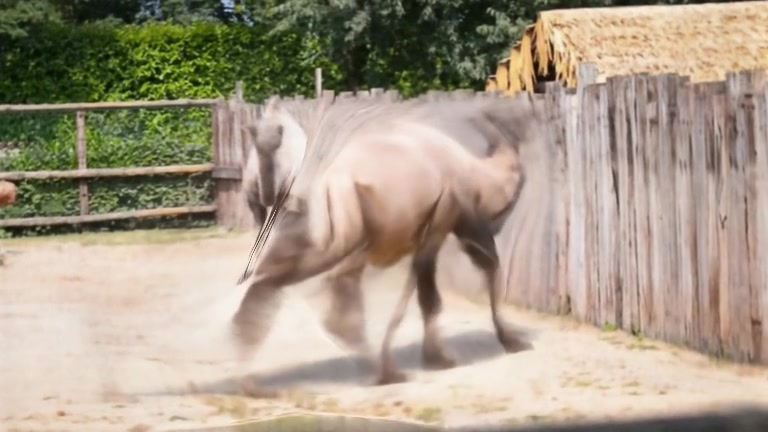}&
\includegraphics[width=0.230\linewidth,height=0.130\linewidth]{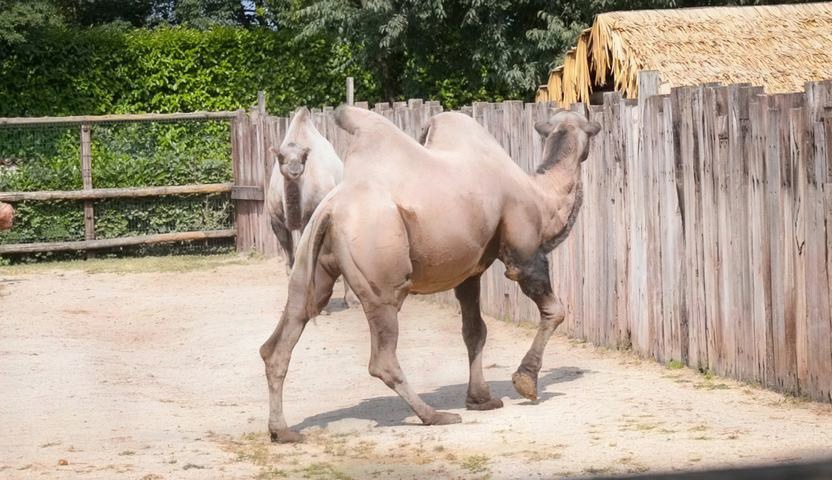}\\

\rotatebox{90}{\small \hspace{2mm}  }&
\includegraphics[width=0.230\linewidth]{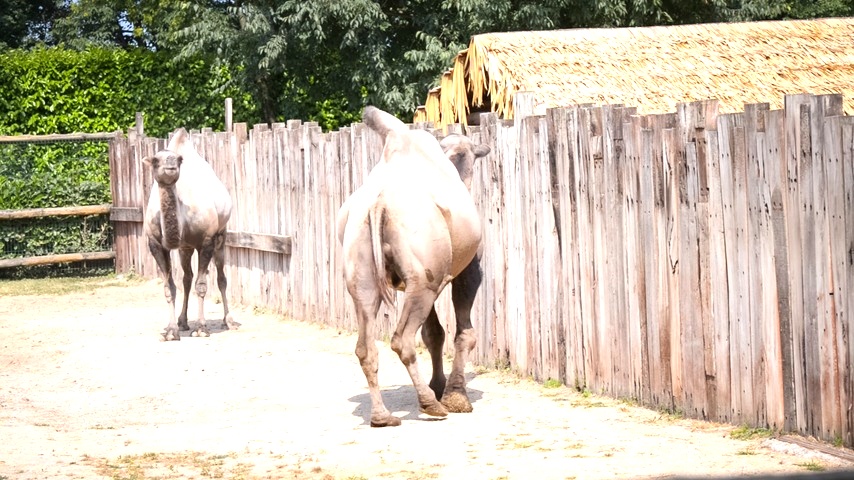}&
\includegraphics[width=0.230\linewidth]
{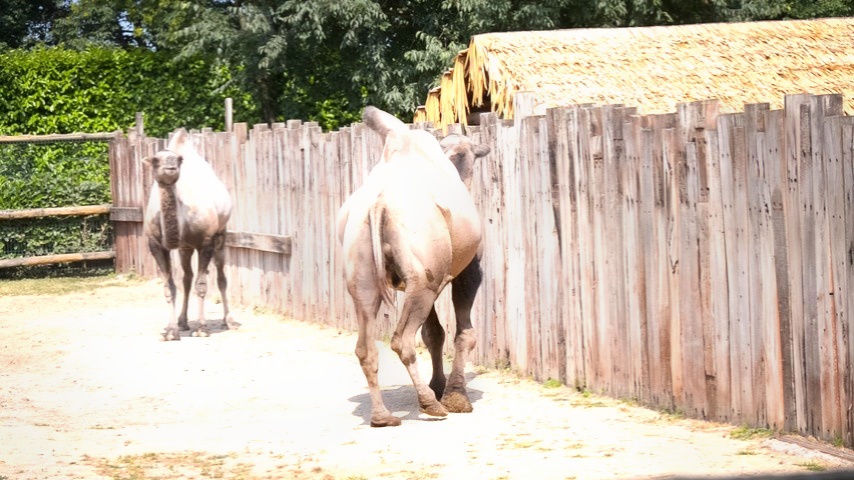}&
\includegraphics[width=0.230\linewidth]{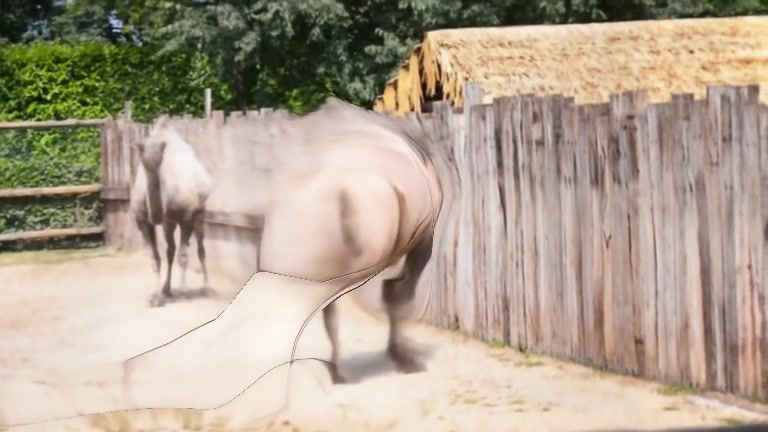}&
\includegraphics[width=0.230\linewidth,height=0.130\linewidth]{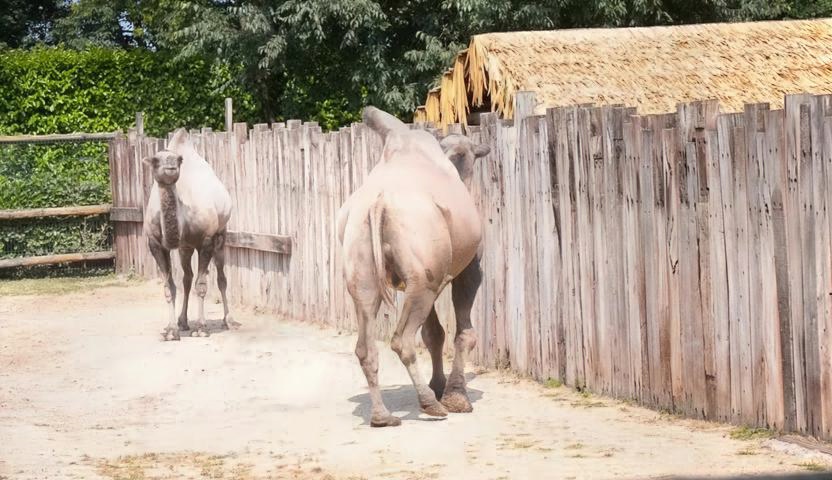}\\

\rotatebox{90}{\small \hspace{2mm}  }&
\includegraphics[width=0.230\linewidth]{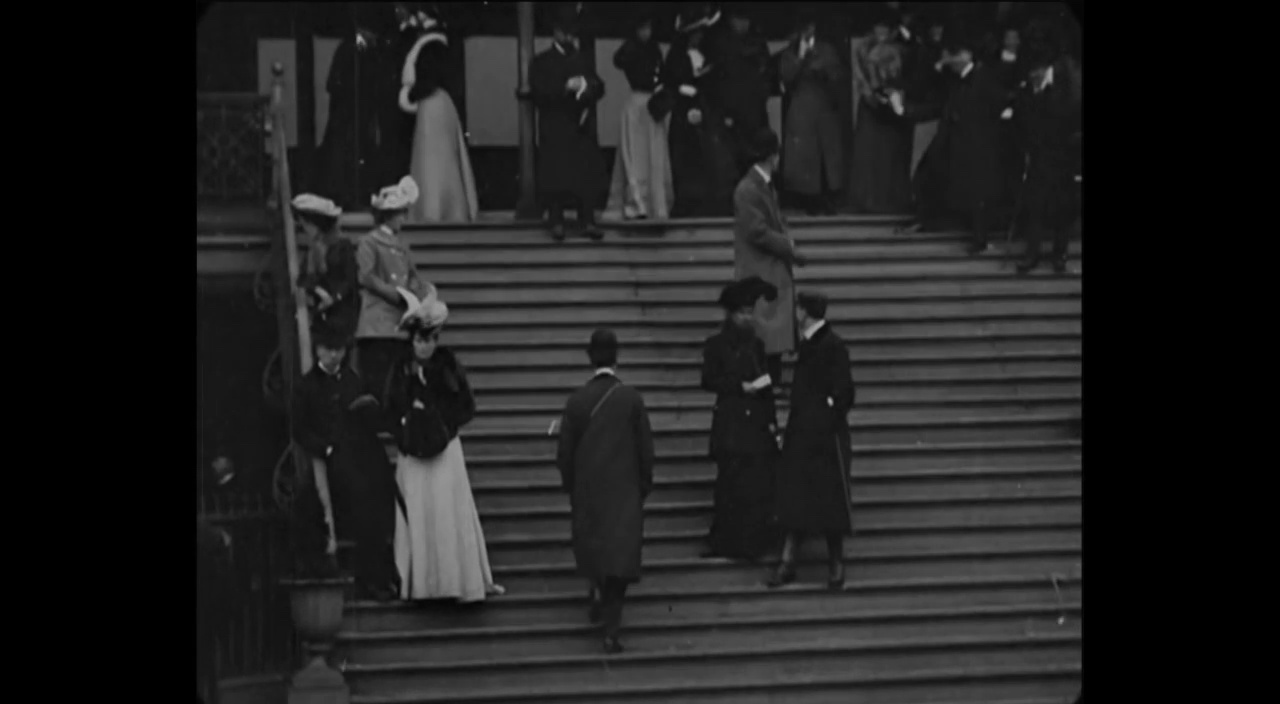}&
\includegraphics[width=0.230\linewidth]{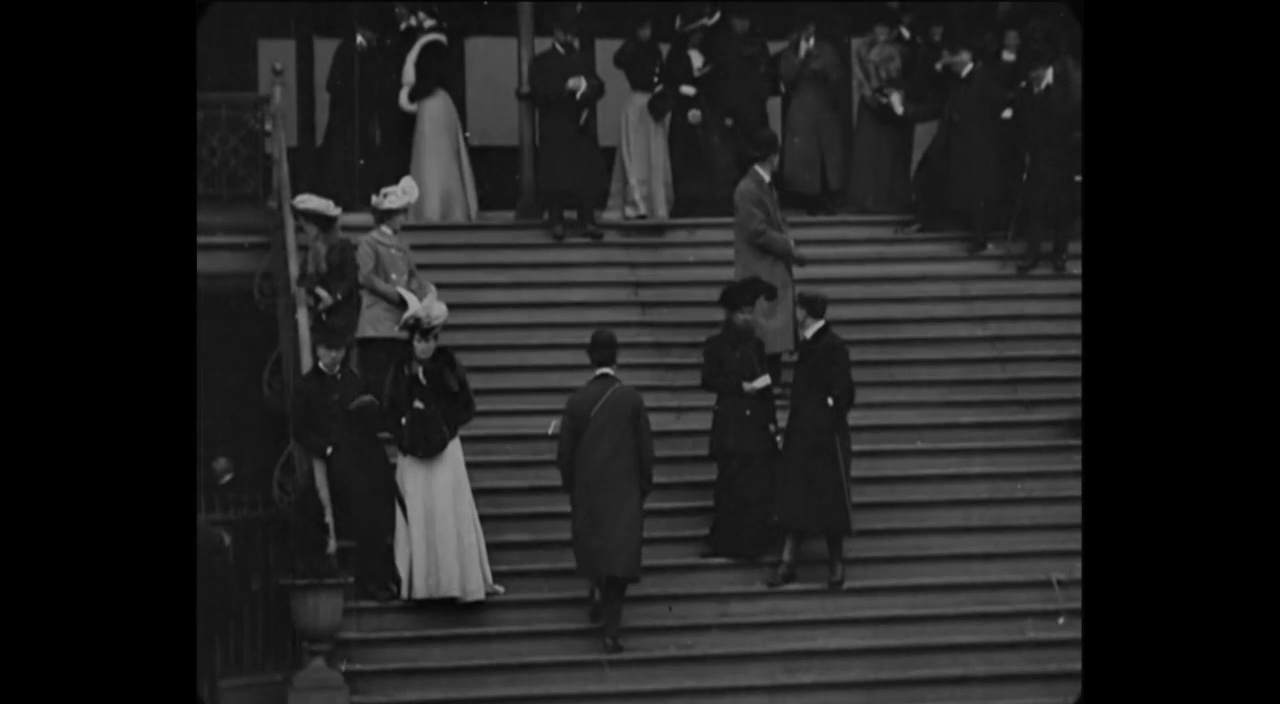}&
\includegraphics[width=0.230\linewidth,height=0.126\linewidth]{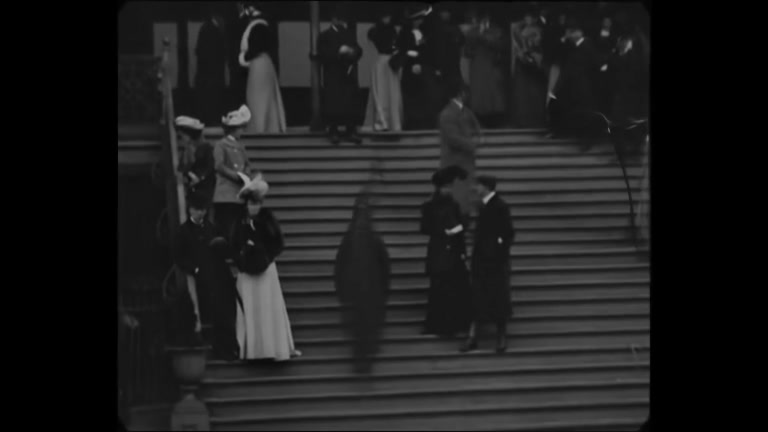}&
\includegraphics[width=0.230\linewidth]{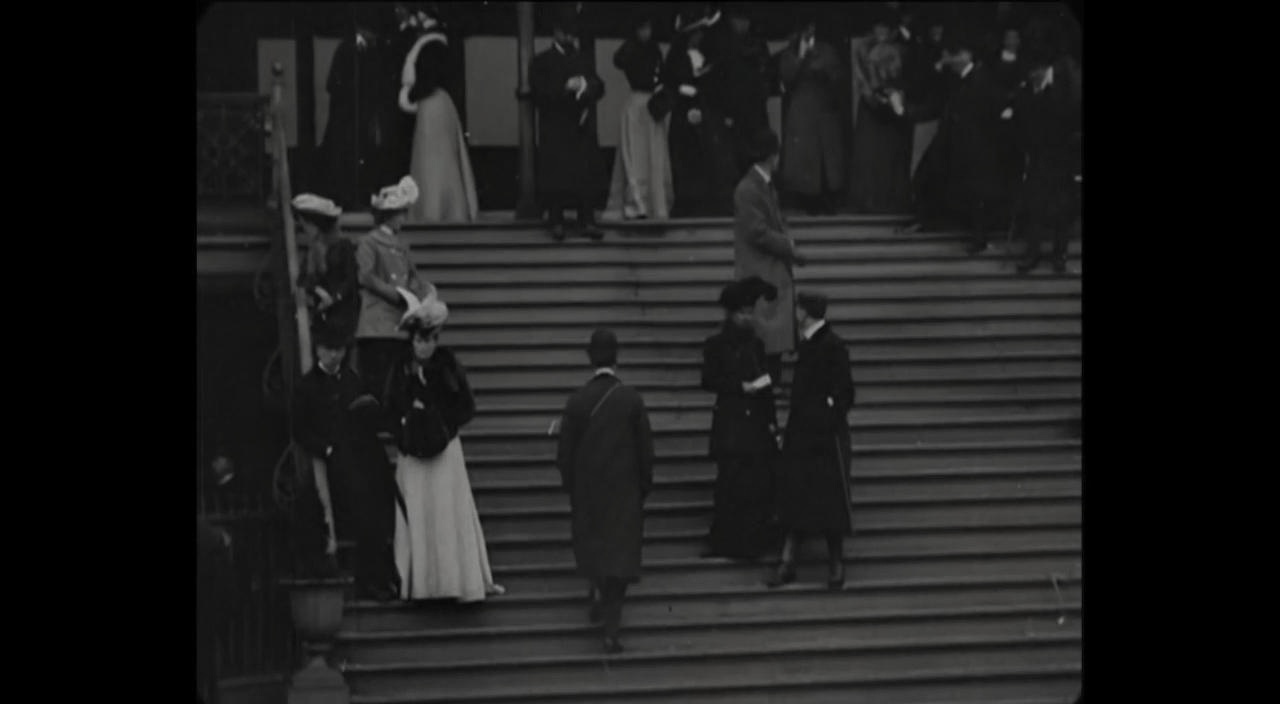}\\

\rotatebox{90}{\small \hspace{2mm}  }&
\includegraphics[width=0.230\linewidth]{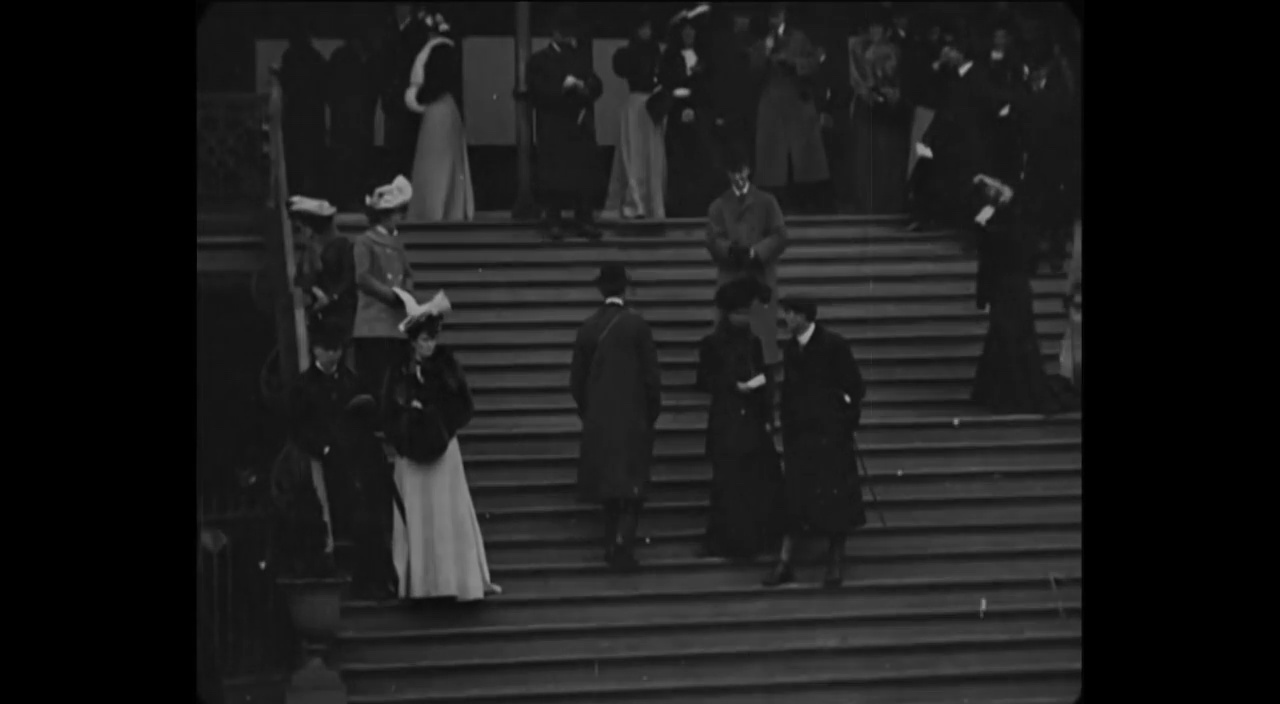}&
\includegraphics[width=0.230\linewidth]{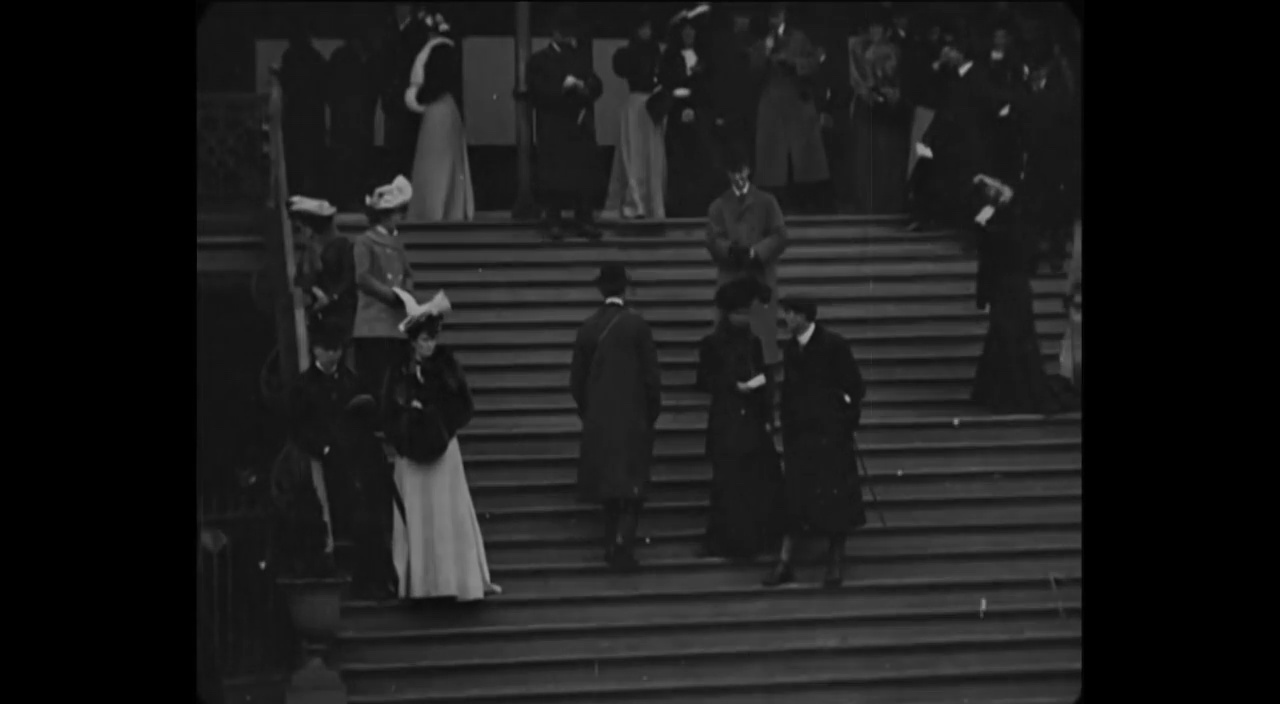}&
\includegraphics[width=0.230\linewidth,height=0.126\linewidth]
{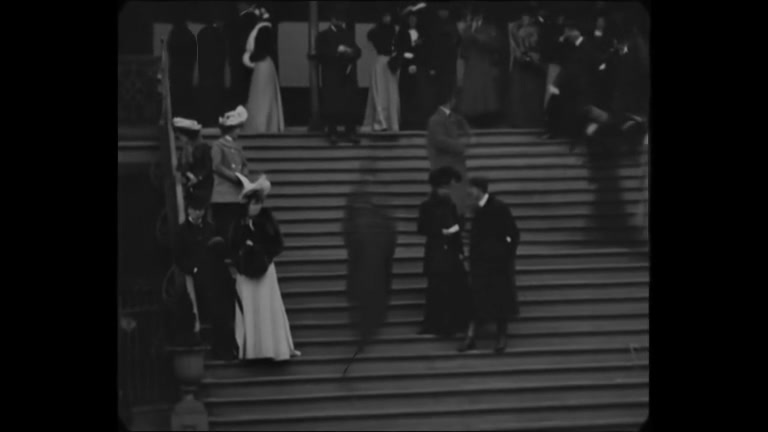}&
\includegraphics[width=0.230\linewidth]{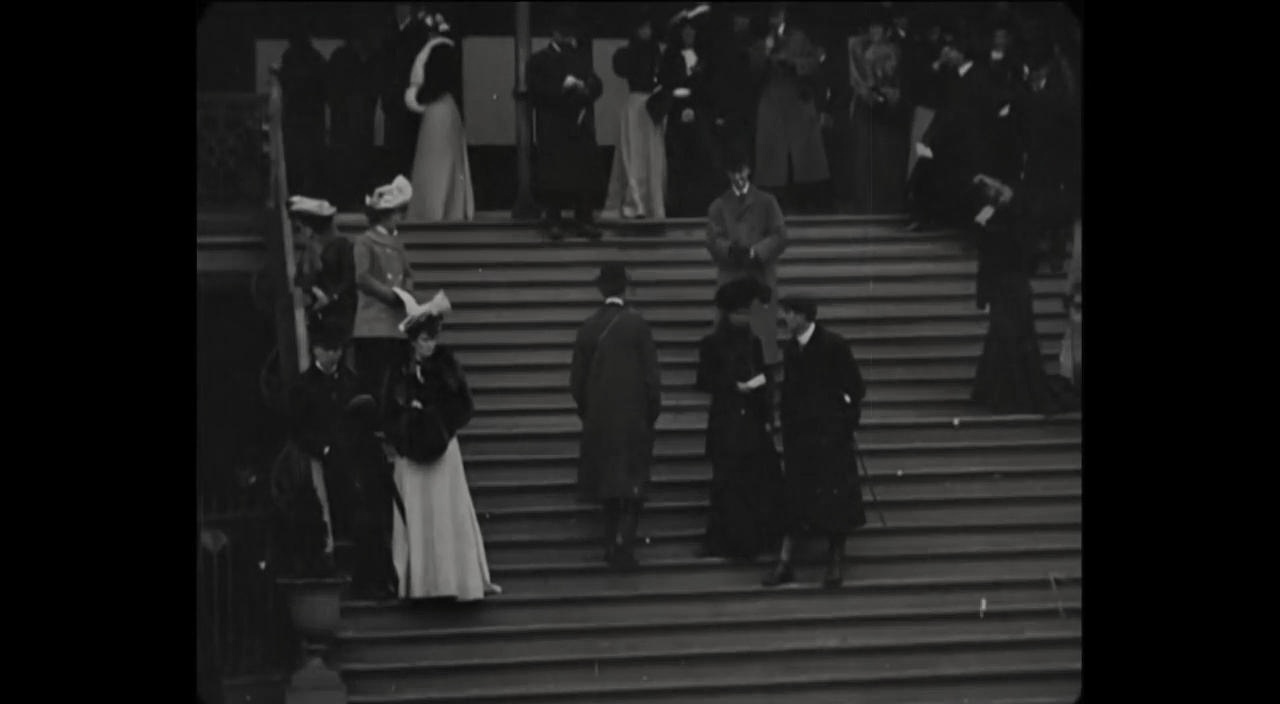}\\

&Input & \small{ConvLSTM}  & \small{Ours atlas only} & \small{Ours} \\
\end{tabular}
\\
\caption{\textbf{Qualitative comparisons to baselines.} Our results outperforms the baseline ConvLSTM significantly on various flickering videos. We highly encourage readers to see videos in our project website.}

\label{fig:baselines}
\vspace{-1em}
\end{figure*}

%% file: sec4_dataset.tex
\section{Blind Deflickering Dataset}
\label{sec:dataset}
%
\begin{table*}[t]
\small
\centering
\renewcommand{\arraystretch}{1.1}
\begin{tabular}{lccccc}
\toprule
\multirow{2}{*}{Task} & \multicolumn{5}{c}{$E_{warp} \downarrow$} \\
 & {Processed} & {Bonneel et al.}~\cite{bonneel2015blind} & Lai et al.~\cite{lai2018learning} & DVP~\cite{DBLP:conf/nips/dvp} &  {Ours} \\ 
\hline
\rowcolor{mygray}
w/o Extra Consistent Guidance & -- & {\color{red} \xmark} & {\color{red} \xmark} & {\color{red} \xmark} &  {\color{green} \cmark} \\
\hline
Dehazing~\cite{he2010single} & 0.120 & 0.128 & 0.136 & \underline{0.109} & \textbf{0.106} \\
Spatial White Balancing~\cite{hsu2008light}  & 0.087 & 0.081 & 0.098 & \underline{0.073} & \textbf{0.062}\\
Colorization~\cite{zhang2016colorful} & 0.109 & \underline{0.096} & {0.100} & {0.097} & \textbf{0.084} \\
Enhancement~\cite{gharbi2017deep} & 0.125 & {0.105} & 0.115 & \underline{0.102} & \textbf{0.093} \\
CycleGAN~\cite{CycleGAN2017} & 0.124 & 0.113 & 0.117 & \underline{0.103} & \textbf{0.099} \\
Intrinsic Decomposition & 0.131 & {0.085} & 0.108 & \underline{0.071} & \textbf{0.069} \\
Style Transfer & 0.202 & {0.161} & 0.177 &  \underline{0.143} & \textbf{0.142} \\
\hline
Average Score & 0.128  & {0.110} & 0.122 & \underline{0.100} & \textbf{0.094} \\
\bottomrule
\end{tabular}
\vspace{-1mm}
\caption{\textbf{Qualitative comparison with blind video temporal consistency methods that use input videos as extra guidance.} While our approach does not use input videos as guidance, our method achieves better numerical performance compared with the baselines. }
\label{table:MainComparison}\
\vspace{-1.6em}
\end{table*}

We construct the first publicly available dataset for blind deflickering. 

\noindent \textbf{Real-world data.}
We first collect real-world videos that contain various types of flickering artifacts. Specifically, we collect five types of real-world flickering videos: 
\begin{itemize}
\setlength{\itemsep}{0pt}
\setlength{\parsep}{0pt}
\setlength{\parskip}{0pt}
    \item \textit{Old movies} contain complicated flickering patterns. The flicker is caused by multiple reasons, such as unstable exposure time and aging of film materials. Hence, the flickering can be high-frequency or low-frequency, globally and locally. 
    \item \textit{Old cartoons} are similar to old movies, but the structures differ from natural videos.
    \item\textit{Time-lapse} videos capture a scene for a long time, and the environment illumination usually changes a lot. 
    \item \textit{Slow-motion} videos can capture high-frequency changes in lighting. 
    \item \textit{Processed} videos denote the videos processed by various processing algorithms. The patterns of flicker are decided by the specific algorithm. We follow the setting in \cite{lai2018learning}. 
\end{itemize}

\begin{figure}[t]
\centering
\begin{tabular}{@{}c@{\hspace{1mm}}c@{\hspace{1mm}}c@{\hspace{1mm}}c@{\hspace{1mm}}c@{\hspace{1mm}}c@{}}

\rotatebox{90}{\small \hspace{0mm} w/o local  refinement}&
\includegraphics[height=0.34\linewidth]{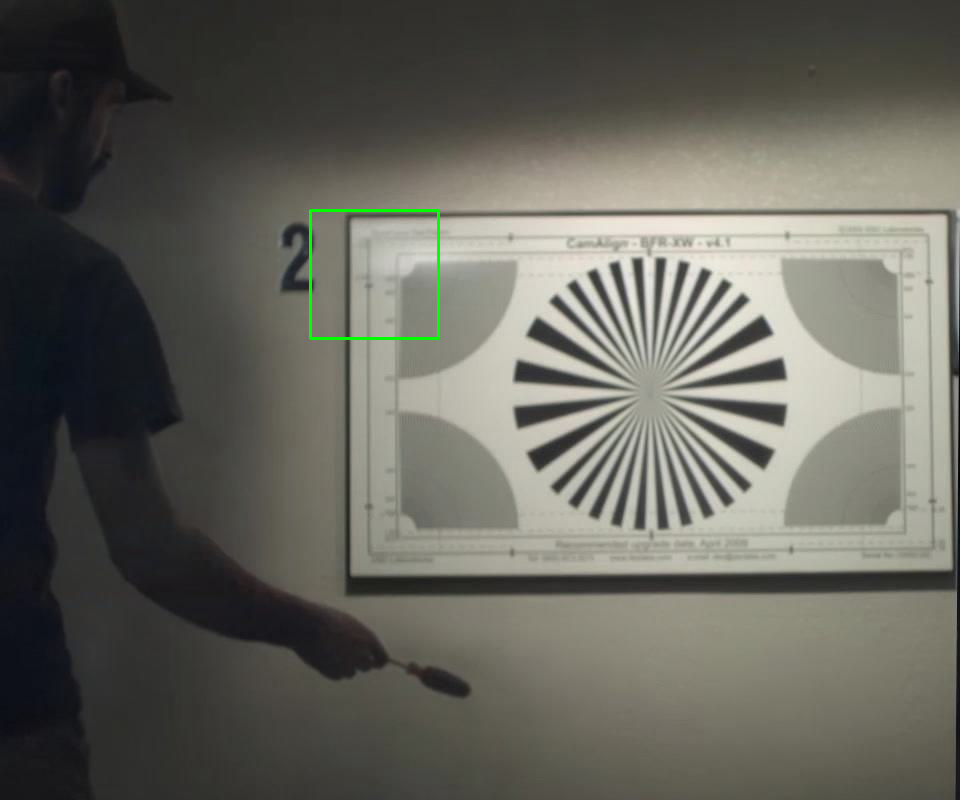}&
\includegraphics[height=0.34\linewidth]{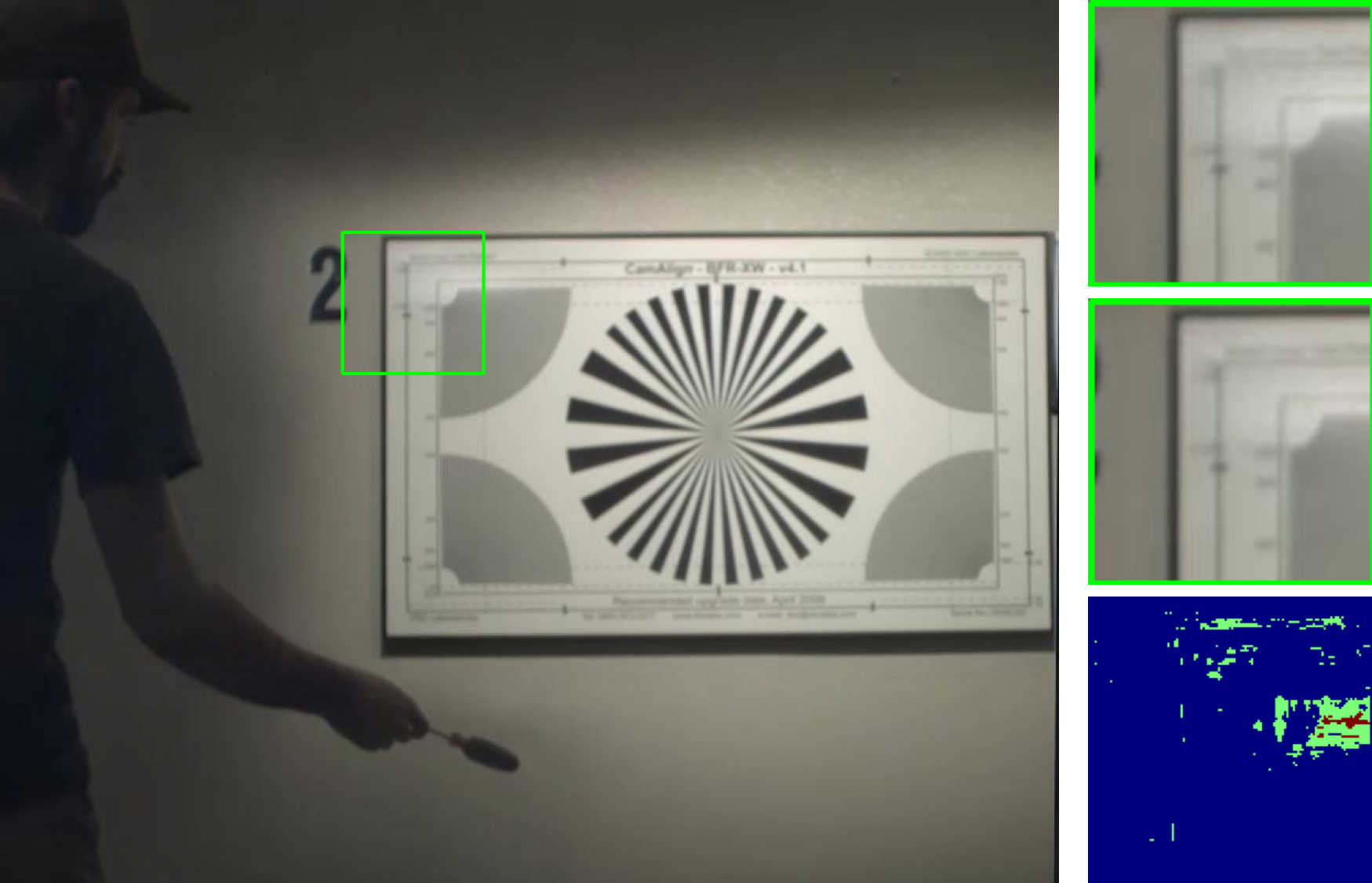}\\
\rotatebox{90}{\small \hspace{7mm} Full model }&
\includegraphics[height=0.34\linewidth]{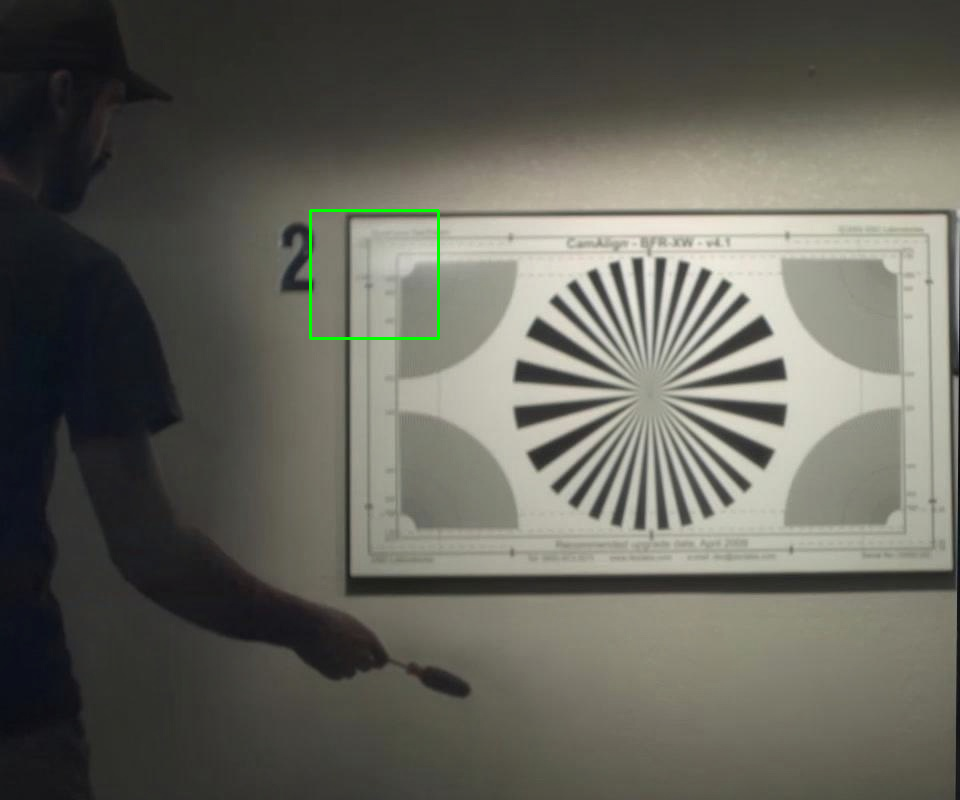}&
\includegraphics[height=0.34\linewidth]{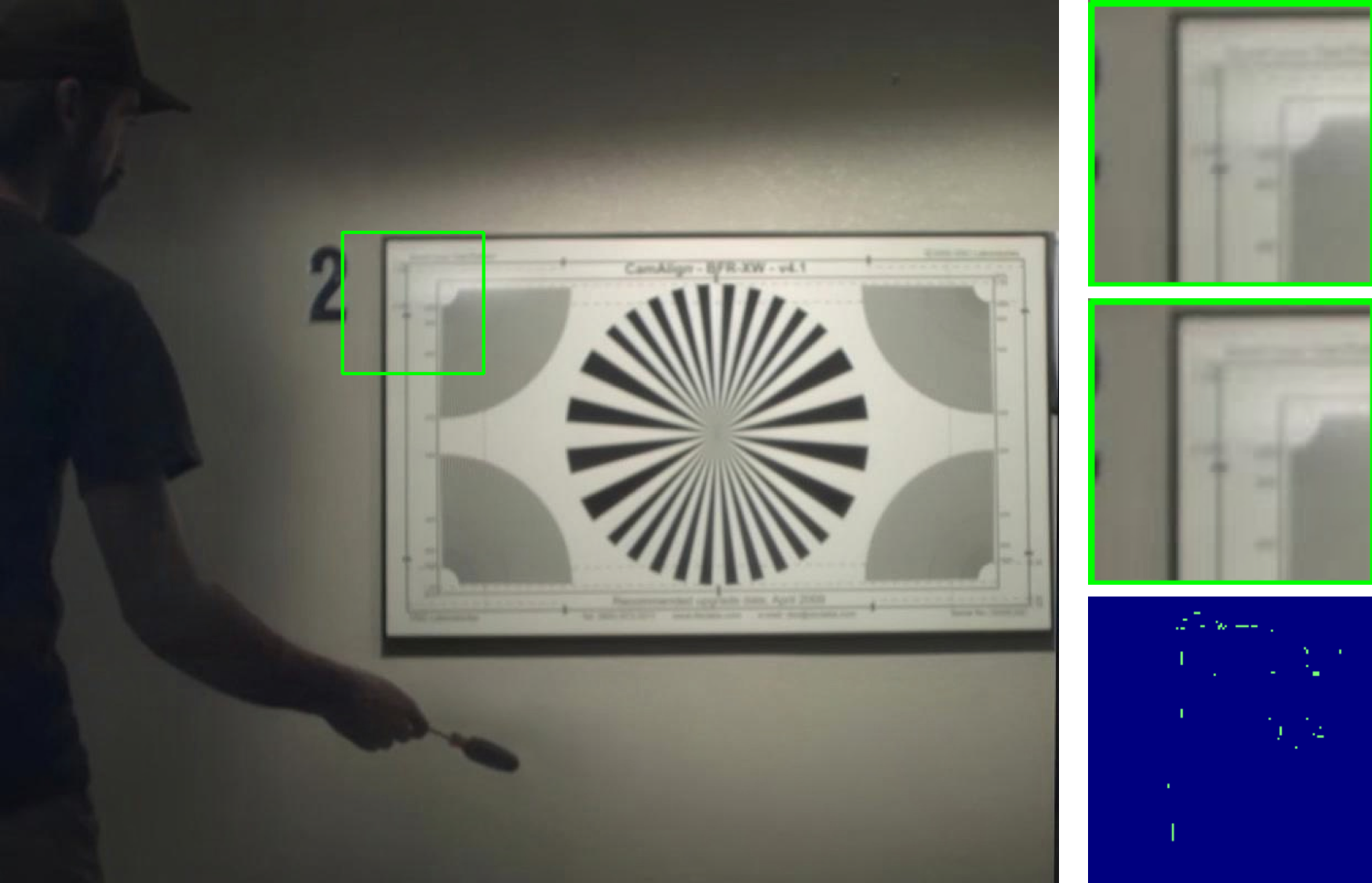}\\
\end{tabular}

\vspace{-0.5em}
\caption{\textbf{Ablation study for local refinement module.} The local refinement network is vital to remove the local flickering. Cropped images and their difference maps are placed at the third column.}
\vspace{-1.1em}
\label{fig:refine}
\end{figure}

\noindent \textbf{Synthetic data.}
While real-world videos are good for evaluating perceptual performance, they do not have ground truth for quantitative evaluation. Hence, we create a synthetic dataset that provides ground truth for quantitative analysis. Let $\{G_t\}_{t=1}^T$ be the clean video frames, the flickered video $\{G_t\}_{t=1}^T$ can be obtained by adding the flickering artifacts $F_t$ for each frame at time $t$:
\begin{align}
    I_t = G_t + F_t, 
\end{align}
where $\{F_t\}_{t=1}^T$ is the synthesized flickering artifacts. For the temporal dimension, we synthesize both short-term and long-term flicker. Specifically, we set a window size $W$, which denotes the number of frames that share the same flickering artifacts. We set the window size $W$ as 1, 3, 10 respectively.

\noindent\textbf{Summary.} We provide $20$, $10$, $10$, $10$, $157$, and $90$ for old movies, old cartoons, slow-motion videos, time-lapse videos, processed videos, and synthetic videos.

%% file: sec5_experiment.tex
\section{Experiments}

\subsection{Evaluation Setup} 

\noindent \textbf{Dataset.} We mainly use our constructed Blind Deflickering Dataset for evaluation. Details are presented in Section~\ref{sec:dataset}.

\noindent \textbf{Evaluation metric.} We measure the temporal inconsistency based on the warping error used in DVP~\cite{DBLP:conf/nips/dvp} that considers both short-term and long-term warping errors for quantitative evaluation. Given a pair of frames $O_t$ and $O_s$, the warping error $E_{pair}$ can be calculated by:
\begin{align}
&E_{pair}(O_t,O_{s}) =  ||M_{t,s} \odot (O_t - \hat{O}_{s})||_1,    \\
&E_{warp}^t=E_{pair}(O_t,O_{t-1}) + E_{pair}(O_t, O_1),
\end{align}
where $\hat{O}_{s}$ is obtained by warping the $O_{s}$ with the optical flow from frame $t$ to frame $s$. $M_{t,s}$ is the occlusion mask from frame $t$ and frame $s$. For each frame $t$, the warping error $E_{warp}^t$ is computed with the previous and first frames in the video. We compute the warping error for all the frames in a video.

\subsection{Comparisons to Baselines}
\noindent \textbf{Baseline.} As our approach is the first method for blind deflickering, no existing public method can be used for comparison. Thus, we design a baseline inspired by blind video temporal consistency approaches: \textit{ConvLSTM}, modified from Lai et al.~\cite{lai2018learning}. Specifically, we replace the consistent input pair of frames with flickered pair frames, and we retrain the ConvLSTM on their training dataset~\cite{lai2018learning}.

\noindent \textbf{Results.} Table~\ref{table:comparison_ourdata}(a) provides quantitative results on two types of videos where we can estimate high-quality optical flow from consistent videos for computing the warping errors. Our results are consistently better than the baseline. In Table~\ref{table:comparison_ourdata}(b), we conduct a user study on Amazon Mechanical Turk following an A/B test protocol~\cite{ChenK17} to evaluate the perceptual preference between the main baseline ConvLSTM and our method. Each user needs to choose a video with better perceptual quality from videos processed by our method and the baseline. We use all real-world videos that cannot obtain high-quality optical flow for quantitative evaluation.
In total, we have $20$ users and $50$ pairs of comparisons.
Our method outperforms the baseline significantly in all tasks. 
Figure~\ref{fig:baselines} shows the qualitative comparisons between our approach and the baseline. Our approach removes various types of flicker, and our results are more temporal consistent than baselines.

\noindent \textbf{Comparisons to blind temporal consistency {methods}.} 
The comparison between our approach and blind video temporal consistency methods is unfair since our approach \textit{does not} require extra videos and baselines \textit{use} extra input videos as guidance. However, we still provide the comparison results for reference. Specifically, we use three state-of-the-art baselines, including Bonneel et al.~\cite{bonneel2015blind}, Lai et al.~\cite{lai2018learning} and DVP~\cite{DBLP:conf/nips/dvp}. Table~\ref{table:MainComparison} shows the quantitative results between our approach and baselines. The warping error of our approach on various processed videos is lower than all the baselines, which indicates that our approach is more temporal consistent even without the input video guidance. 
\begin{table}[t]
\begin{center}
\resizebox{0.8\linewidth}{!}{
\begin{tabular}{l@{\hspace{3em}}c}
\toprule
\multicolumn{1}{l}{Method} & $E_{warp} \downarrow$  \\
\midrule
Ours without atlas \& neural filtering & 0.131  \\
Ours without local refinement & 0.090  \\
Ours full model & \textbf{{0.088}} \\
\bottomrule
\end{tabular}
}
\end{center}
\vspace{-1.2em}
\caption{ \textbf{Quantitative results of ablation study}. The atlas and neural filtering strategy reduce temporal inconsistency significantly. The local refinement makes a slight difference in warping error but does improve perceptual performance.}
\label{table:ablation}
\vspace{-1em}
\end{table}

\subsection{Ablation Study}

We present the flawed atlas in the third column of  {Figure~\ref{fig:baselines}}. While the temporal consistency between frames is perfect, the atlas-based frames contain many artifacts and distortions. Hence, designing dedicated strategies is essential for blind deflickering. In this part, we analyze the importance of our two modules, respectively: (1) \textit{neural filtering} denotes the atlas generation and neural filtering since they cannot be split. (2) \textit{local refinement} denotes the local refinement module.

\noindent \textbf{Neural filtering.} We first analyze the importance of neural filtering by removing this part. We direct apply our local refinement module on the flickering input videos. As shown in Table~\ref{table:ablation}, removing the neural filtering module significantly increases temporal inconsistency. 

\noindent \textbf{Local refinement.} As shown in Table~\ref{table:ablation}, removing the local refinement network degraded the quantitative performance slightly. In Figure~\ref{fig:refine}, we show some local regions in the frames are temporal inconsistent. While these regions are small and the warping errors are only slightly reduced, this local flickering does hurt the perceptual performance, as shown in our supplementary materials.

\begin{figure}[t]
\centering
\begin{tabular}{@{}c@{\hspace{1mm}}c@{\hspace{1mm}}c@{}}

\includegraphics[width=0.320\linewidth]{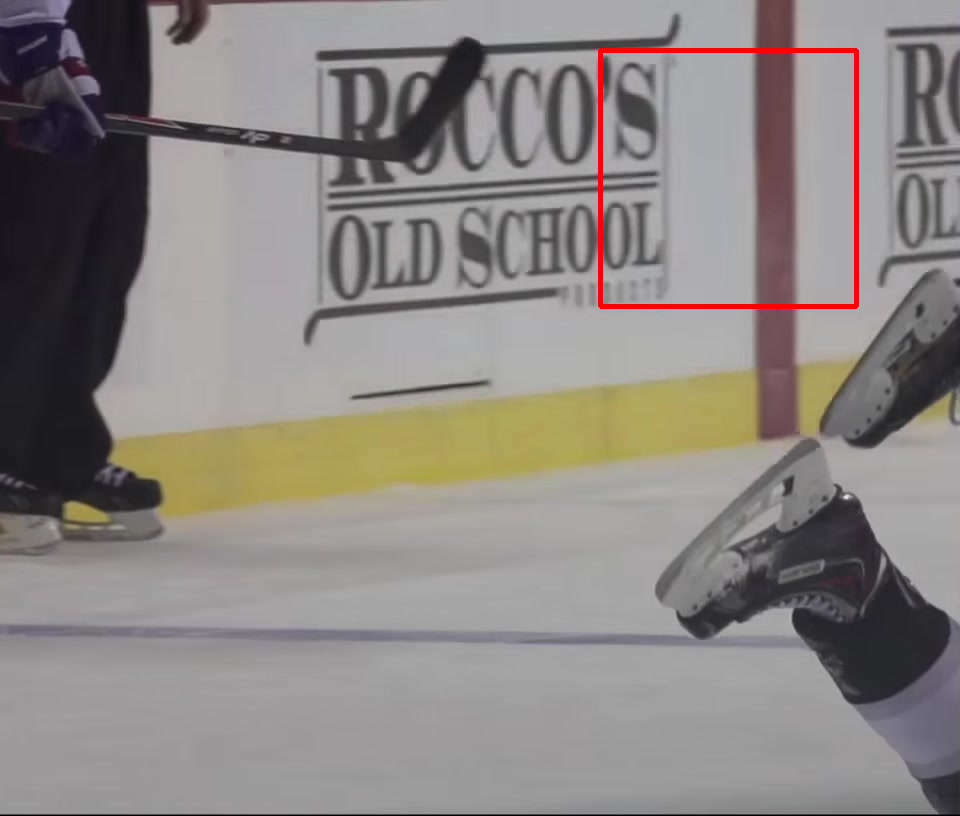}&
\includegraphics[width=0.320\linewidth]{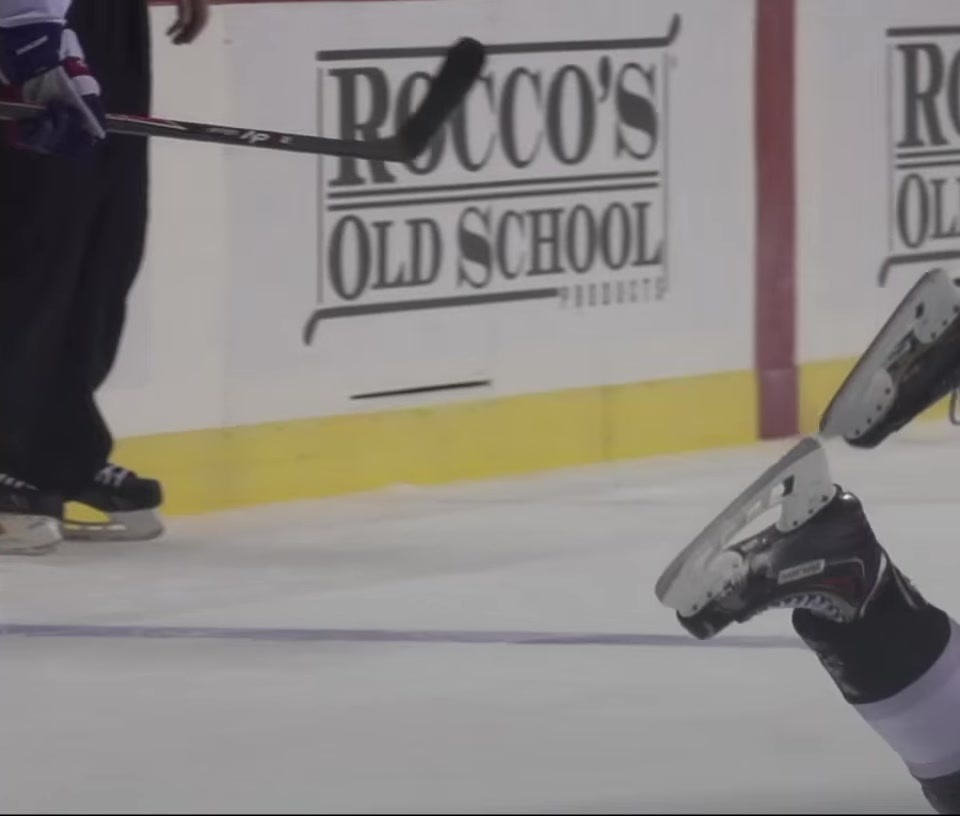}&
\includegraphics[width=0.320\linewidth]{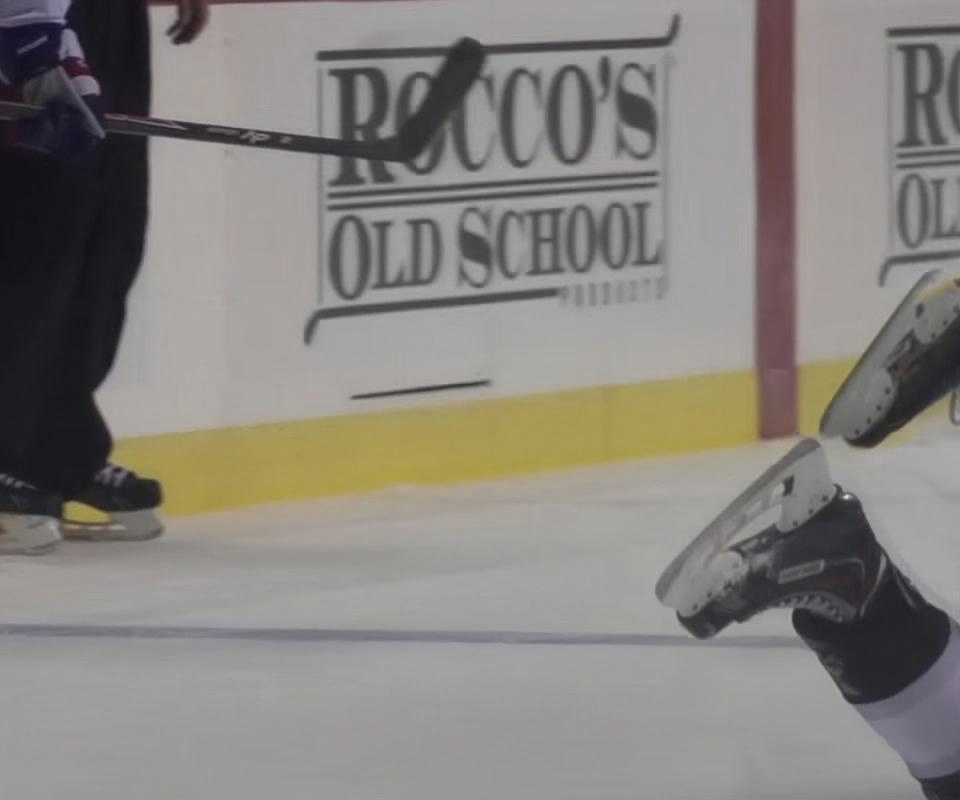} \\

\includegraphics[width=0.320\linewidth]{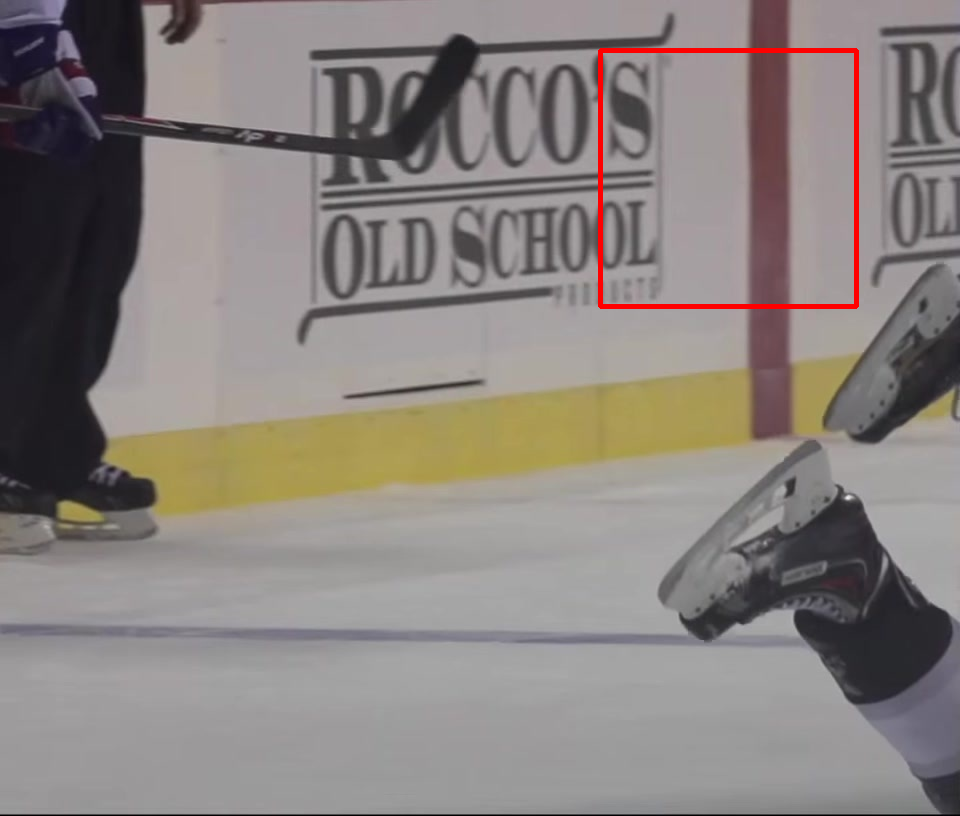}&
\includegraphics[width=0.320\linewidth]{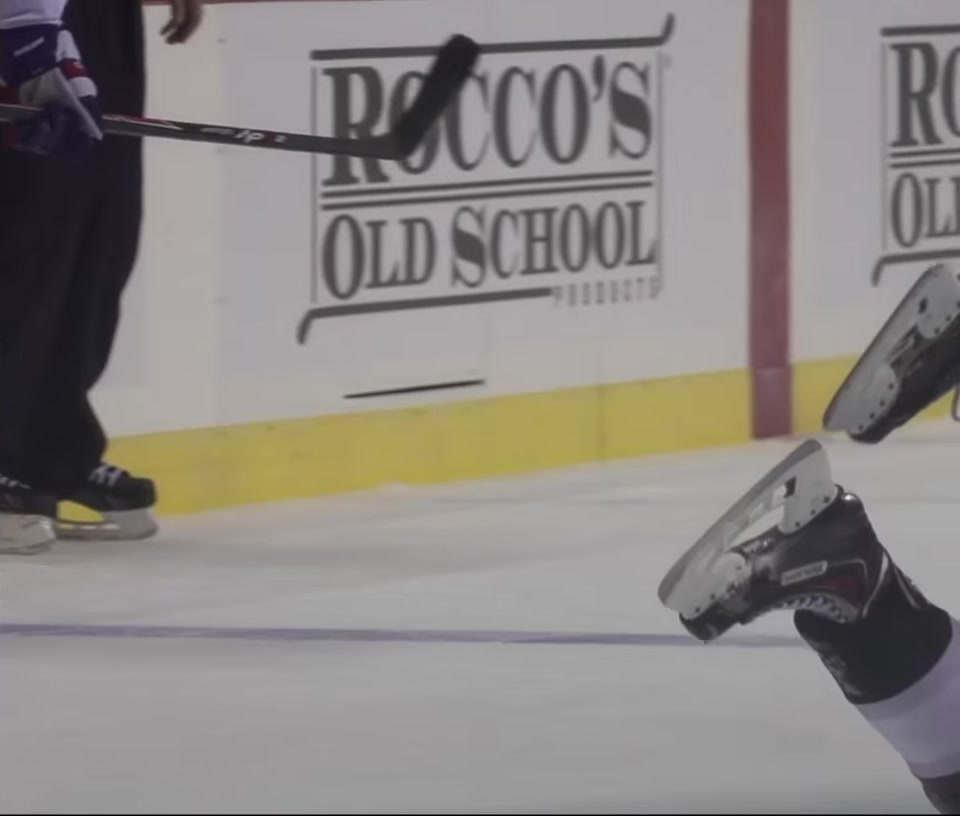}&
\includegraphics[width=0.320\linewidth]{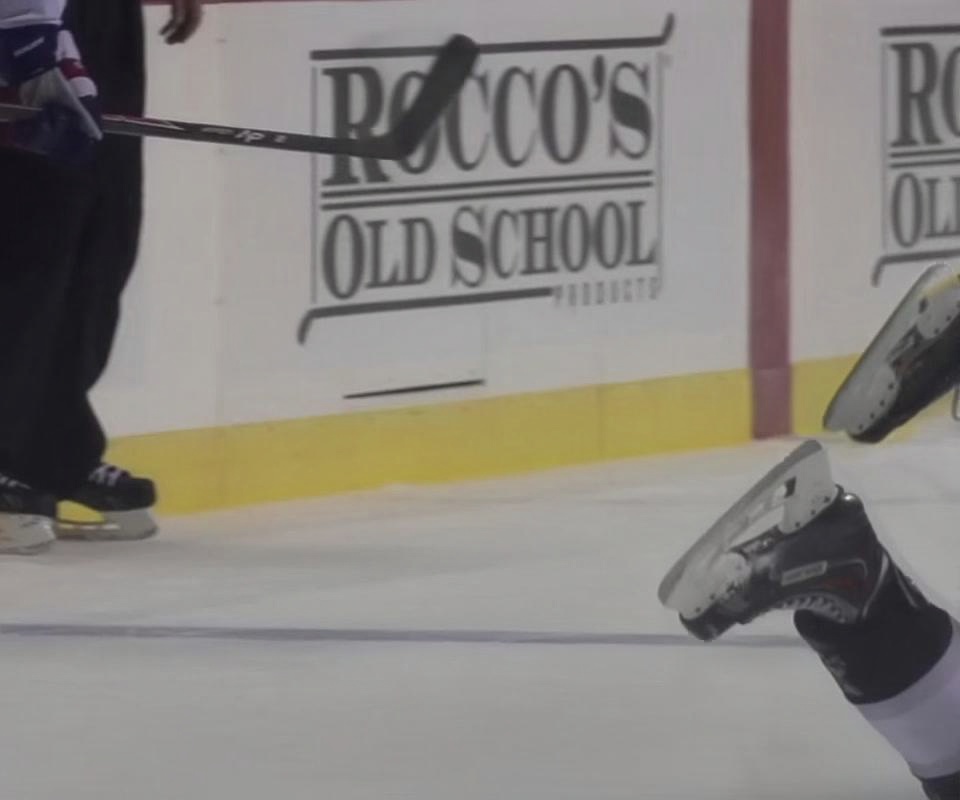} \\
Input & Human experts & Ours \\

\end{tabular}
\vspace{-0.5em}
\caption{\textbf{Comparison to human experts}.
Input: the lower frame is redder than the upper one. Our method achieves comparable performance with human experts in this case. }
\label{fig:exper}
\vspace{-1.2em}
\end{figure}

\subsection{Comparisons to Human Experts}

We compare our approach to human experts that use commercial software for deflickering. We adopt the RE:Vision DE:Flicker commercial software~\cite{revision} for comparison, following Bonneel et al.~\cite{bonneel2015blind}. Specifically, we obtain the official demos processed by experts and compare them. Figure~\ref{fig:exper} shows the qualitative comparison results. We can see that our approach can obtain competitive results in a fully-automatic manner. Besides, as discussed in Bonneel et al.~\cite{bonneel2015blind}, the videos processed by new users are usually low-quality. 

\subsection{Discussion and Future Work}
\noindent\textbf{Potential applications.} Our model can be applied to all evaluated types of flickering videos. Besides, while our approach is designed for videos, it is possible to apply \textit{Blind Deflickering} for other tasks (e.g., novel view synthesis~\cite{nerf,xie2022high}) where flickering artifacts exist. 

\noindent\textbf{Temporal consistency beyond our scope.} Solving the temporal inconsistency of video content is beyond the scope of deflickering. For example, the contents obtained by video generation algorithms can be very different. Large scratches in old films can destroy the contents and result in unstable videos, which require extra restoration technique~\cite{wan2022bringing}. We leave the study for removing these temporally inconsistent artifacts to the future work.